\documentclass[journal,twoside,web]{ieeecolor}
\usepackage{generic}

\usepackage{cite}
\usepackage{amsmath,amssymb,amsfonts}
\usepackage{graphicx}
\usepackage{textcomp}
\def\BibTeX{{\rm B\kern-.05em{\sc i\kern-.025em b}\kern-.08em
    T\kern-.1667em\lower.7ex\hbox{E}\kern-.125emX}}
\usepackage{comment}
\usepackage{bm}

\usepackage[utf8]{inputenc} 
\usepackage[T1]{fontenc}    
\usepackage{url}            
\usepackage{booktabs}       
\usepackage{diagbox}        
\usepackage{makecell}
\usepackage{nicefrac}       
\usepackage{microtype}      
\usepackage{xspace}
\usepackage{bm}
\usepackage{times}
\usepackage[ruled,vlined]{algorithm2e}
\definecolor{customblue}{rgb}{0.36, 0.55, 0.75}
\usepackage[colorlinks=true,linkcolor=magenta, urlcolor=magenta,citecolor=customblue, anchorcolor=magenta]{hyperref}
\definecolor{myred}{RGB}{220,0,0}
\definecolor{myblue}{RGB}{0,80,255}
\definecolor{mygreen}{RGB}{0,150,0}
\usepackage{marvosym}
\usepackage{booktabs}
\usepackage{tabularx}
\usepackage{multirow}
\newcommand{\revise}[1]{{\color{blue}#1}}

\usepackage{wrapfig}

\usepackage{bbm}
\usepackage{multirow}
\usepackage{float}
\usepackage{subcaption}
\usepackage{scalerel}

\usepackage{array}
\newcolumntype{H}{>{\setbox0=\hbox\bgroup}c<{\egroup}@{}}

\newcommand{\cut}[1]{}  
\newcommand{\std}[1]{\scriptsize{$\pm$#1}}

\begin{document}

\title{Benchmarking ERP Analysis: Manual Features, Deep Learning, and Foundation Models
}

\author{
Yihe Wang,
Zhiqiao Kang,$^*$
Bohan Chen,$^*$
Yu Zhang,
Xiang Zhang\textsuperscript{\Letter} 
\thanks{Yihe Wang and Xiang Zhang are with the Department of Computer Science,  University of North Carolina-Charlotte, Charlotte, North Carolina 28223, United States. Xiang Zhang is the corresponding author (E-mail: xiang.zhang@charlotte.edu).}
\thanks{Zhiqiao Kang is with the College of Future Technology, South China University of Technology, Guangzhou, Guangdong 511442, China.}
\thanks{Bohan Chen is with the Department of Statistics, University of California-Davis, Davis, California 95618, United States.}
\thanks{Yu Zhang is with the Department of Psychiatry and Behavioral Sciences, Stanford University School of Medicine, Wu Tsai Neurosciences Institute, and Stanford Institute for Human-Centered Artificial Intelligence, Stanford, California 94305, United States.}
}

\maketitle
\def\thefootnote{*}\footnotetext{Equal Contribution. This work was conducted during their research internship at UNC Charlotte.}

\begin{abstract}
Event-related potential (ERP), a specialized paradigm of electroencephalographic (EEG), reflects neurological responses to external stimuli or events, generally associated with the brain’s processing of specific cognitive tasks. ERP plays a critical role in cognitive analysis, the detection of neurological diseases, and the assessment of psychological states. Recent years have seen substantial advances in deep learning-based methods for spontaneous EEG and other non-time-locked task-related EEG signals. However, their effectiveness on ERP data remains underexplored, and many existing ERP studies still rely heavily on manually extracted features. In this paper, we conduct a comprehensive benchmark study that systematically compares traditional manual features (followed by a linear classifier), deep learning models, and pre-trained EEG foundation models for ERP analysis. We establish a unified data preprocessing and training pipeline and evaluate these approaches on two representative tasks, ERP stimulus classification and ERP-based brain disease detection, across 12 publicly available datasets. Furthermore, we investigate various token-embedding strategies within advanced Transformer architectures to identify embedding designs that better suit ERP data. Our study provides a landmark framework to guide method selection and tailored model design for future ERP analysis. The code is available at \url{https://github.com/DL4mHealth/ERP-Benchmark}

\end{abstract}

\begin{IEEEkeywords}
ERP, event-related potential, EEG, brain-computer interfaces, deep learning, foundation model
\end{IEEEkeywords}

\section{Introduction}
\label{sec:intro}

\begin{figure}[t]
    \centering \includegraphics[width=1.0\columnwidth]{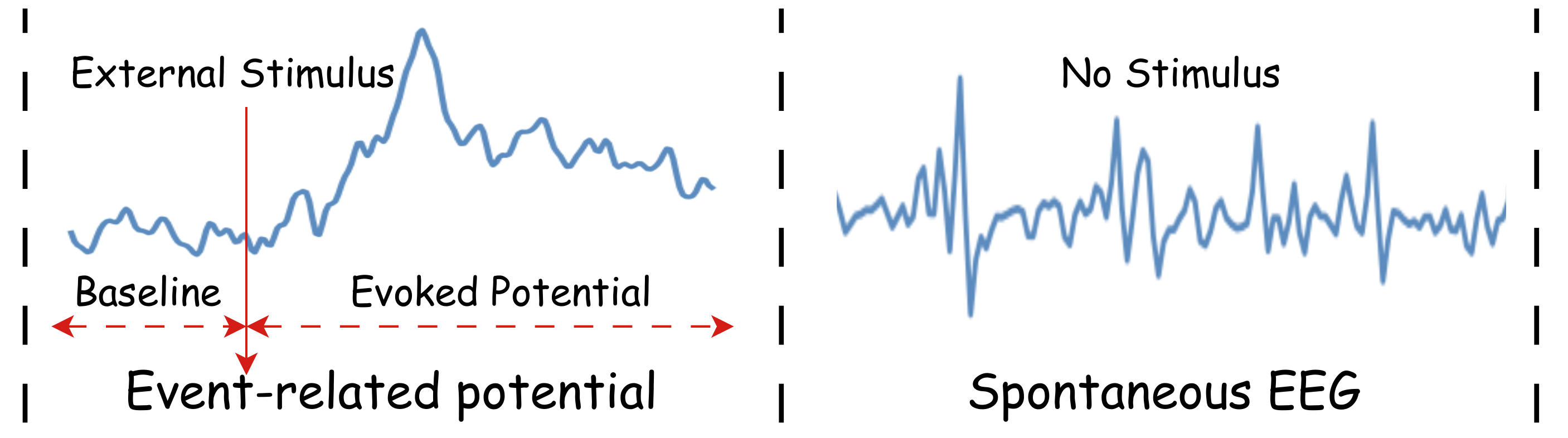}
    \caption{\textbf{ERP vs Spontaneous EEG.} ERP is inherently different from Spontaneous EEG, requiring specific analysis methods.
    }
    \label{fig:erp_eeg_comparison}
    \vspace{-6mm}
\end{figure}

ERP is an EEG-derived measure of time-locked neural activity, serving as a foundational tool with broad applications in neuroscience, dementia, and cognitive science~\cite{macnamara2022event, pruvost2022evoked, allison2022recording}. As a specialized category within the broader EEG signal domain, ERP exhibits distinct characteristics compared to spontaneous EEG activity~\cite{light2010electroencephalography}. While spontaneous EEG reflects ongoing intrinsic brain states but with limited functional specificity, ERP captures time-locked neurological responses elicited by specific sensory, audio, or cognitive events~\cite {luck2012event, helfrich2019cognitive, deboer2013methods}. By aligning neural signals to controlled stimuli, ERP provides a direct and interpretable window into transient cognitive processes. The primary advantage of ERP lies in its temporal synchrony between external events and internal brain electrical responses, which typically yields a higher signal-to-noise ratio than spontaneous EEG activity~\cite{zhang2021survey}. In practice, ERP trials are extracted from continuous EEG recordings by segmenting event-centered epochs~\cite{allison2022recording}. A short pre-stimulus interval (typically 0.2-0.5 seconds) is used for baseline correction to suppress background noise, followed by a longer post-stimulus interval (typically 0.8-1.5 seconds) that contains the stimulus-evoked neural potential~\cite{sur2009event}. Figure~\ref{fig:erp_eeg_comparison} illustrates a signal comparison between ERP and spontaneous EEG.

In recent years, deep learning has achieved substantial progress in EEG decoding across a wide range of applications, including motor imagery control~\cite{mathiyazhagan2025motor}, emotion recognition~\cite{fang2025neuript}, sleep stage classification~\cite{fang2025neuript}, seizure detection~\cite{zheng2025fapex}, and dementia detection~\cite{wang2024adformer}. In addition, advanced deep learning techniques, such as foundation models that have demonstrated remarkable success in domains like natural language processing (NLP) and computer vision (CV) (e.g., ChatGPT~\cite{achiam2023gpt}, Gemini~\cite{team2024gemini}, and DINOv2~\cite{oquab2023dinov2}), are also being explored for learning generalizable EEG representations. Recent EEG foundation models, including LaBraM~\cite{jiang2024large}, EEGPT~\cite{wang2024eegpt}, and CBraMod~\cite{wang2024cbramod}, have reported strong performance on various EEG downstream tasks. However, most existing deep learning model designs primarily target spontaneous EEG signals, such as resting-state and sleep EEG~\cite{fang2025neuript}, or non-time-locked task-induced EEG, such as motor imagery~\cite{hameed2025enhancing}. In contrast, time-locked evoked EEG signals, exemplified by ERP, remain significantly underexplored. As a result, ERP analysis continues to rely mainly on manual feature extraction and conventional statistical approaches, or directly applies models designed for spontaneous EEG on ERP analysis, resulting in suboptimal performance. Given the distinct signal characteristics of ERP and its fundamental differences from spontaneous EEG, it remains unclear to what extent recent advances in deep learning are directly applicable to ERP decoding. This gap necessitates systematic benchmarking of methods for ERP analysis.

To this end, we conduct a comprehensive and fair benchmark evaluation of ERP classification methods. We implement a wide range of approaches, including 2 manual feature extraction methods (with 31 and 91 handcrafted features, respectively), 10 state-of-the-art deep learning methods for EEG, and 3 advanced EEG foundation models, within a unified data preprocessing and evaluation framework. The benchmark covers two representative and practical ERP classification tasks across \textbf{12 public datasets}. The first focuses on \textbf{ERP stimulus classification}, such as distinguishing standard and target stimuli in oddball paradigms, to evaluate models' ability to capture fundamental stimulus-locked cognitive responses. The second addresses \textbf{ERP-based brain disease detection}, such as identifying neurological or psychiatric conditions, thereby assessing model utility in more complex and clinically relevant scenarios. Furthermore, to guide the future development of more effective models, we conduct a comparative analysis of three prevalent token-embedding strategies for EEG data used in recent Transformer architectures. This embedding comparison aims to identify the optimal embedding method for ERP classification, offering critical insights for designing dedicated Transformer backbones for ERP analysis. All the tasks are performed under a rigorously \textbf{subject-independent} setup to evaluate cross-subject learning ability.

In summary, our motivation is to answer the following question through this systematic benchmark study:

\begin{itemize}
    \item \textbf{Q1:} How do deep learning-based EEG methods compare with traditional manual feature extraction methods when applied to ERP data?
    \item \textbf{Q2:} Do EEG Foundation Models with pre-trained weights outperform supervised EEG deep learning models trained from scratch on ERP classification tasks?
    \item \textbf{Q3:} What is the most robust and generalizable method for ERP classification so far?
    \item \textbf{Q4:} Which token embedding strategy is most suitable for designing ERP-specific Transformer models?
\end{itemize}

To provide a concise \textbf{takeaway message}, we summarize the answers below for the reader’s convenience:
\begin{itemize}
    \item \textbf{A1:} Most deep learning methods consistently outperform manual feature extraction approaches on ERP tasks.
    \item \textbf{A2:} Existing EEG foundation models do not demonstrate clear performance advantages over supervised deep learning models trained from scratch.
    \item \textbf{A3:} \textbf{EEGConformer} achieves the best average ranking and shows the most competitive overall performance.
    \item \textbf{A4:} Uni-variate token embedding (See Figure~\ref{fig:token_embedding}) achieves better overall performance than multi-variate and whole-variate patching strategies for ERP-specific Transformer.
\end{itemize}

\section{Related Work}
\label{sec:related}

\textbf{Manual Feature Extraction:} There is a long history of identifying potential biomarkers in ERP signals for classification or brain interpretation. Different types of features are used, including statistical features like mean, skewness, kurtosis, and standard deviation~\cite{tzimourta2019eeg, tzimourta2019analysis, kulkarni2017extracting}, time-domain shape features such as line length, peak-to-peak amplitude, and positive/negative peak amplitudes~\cite{mak2012eeg, li2013identifying, wallace2017eeg}, spectral distribution features like phase shift, phase coherence, bispectrum, and bicoherence~\cite{fraga2013characterizing, tait2019network, waser2016quantifying, trambaiolli2011improving}, frequency-band power features like power spectrum density, relative band power, ratio of EEG rhythm, and energy~\cite{schmidt2013index, liu2016multiple, kanda2014clinician}, complexity features like Shannon entropy, Tsallis entropy, and permutation entropy~\cite{garn2015quantitative, azami2019multiscale, tylova2018unbiased}, and Hjorth parameters such as activity, mobility, and complexity~\cite{peng2021early, mehmood2022eeg, deng2025diagnosis}.

\begin{table*}[t]
    \centering
    \caption{\textbf{Dataset Statistics.} All the datasets follow the same preprocessing pipeline, including a band-pass filter, artifact removal, and resampling to 200Hz. The epoch and baseline window depend on the ERP tasks and datasets. Abbreviations: \textbf{AODD}: Auditory Oddball; \textbf{VODD}: Visual Oddball; \textbf{MSIT+}: Extended multi-source interference task; \textbf{SIM}: Simon Conflict; \textbf{RL}: Reinforcement Learning; \textbf{HC}: Healthy Control; \textbf{PD}: Parkinson's Disease; \textbf{ADHD}: Attention Deficit Hyperactivity Disorder.
    }
    \vspace{-2mm}
    \label{tab:processed_data}
    \resizebox{1.0\textwidth}{!}{%
    \begin{tabular}{@{}l|ccccccc@{}}
    \toprule
    \multicolumn{1}{l|}{\textbf{Datasets}}  & \textbf{ERP Task} & \textbf{\#Subjects} &  \textbf{Baseline(s)} &  \textbf{Epoch(s)}  & \textbf{\#Trials} & \textbf{\#Channels} & \textbf{Classes}  \\ 
    \midrule
    \multicolumn{1}{l|}{\textbf{CESCA-AODD}} & AODD & 127 &  [-0.2, 0]  & [-0.2, 0.8] & 38,151 & 26 & Standard vs Target \\
    \multicolumn{1}{l|}{\textbf{CESCA-VODD}} & VODD & 127 &  [-0.2, 0]  & [-0.2, 0.8] & 20,419 & 26 & Standard vs Target \\
    \multicolumn{1}{l|}{\textbf{CESCA-FLANKER}} & Flanker & 73 &  [-0.2, 0]  & [-0.2, 0.8] & 29,774 & 26 &  Congruent vs Incongruent\\
    \multicolumn{1}{l|}{\textbf{mTBI-ODD}} & AODD & 96 &  [-0.2, 0]  & [-0.2, 0.8] & 24,885 & 61 & Standard vs Target vs Novel \\
    \multicolumn{1}{l|}{\textbf{NSERP-MSIT}} & MSIT+ & 42 &  [-0.5, 0]  & [-0.5, 1.0] & 16,729 & 123 & \makecell{Non-Conflict vs Simon Effect vs \\ Flanker Effect vs Double-Conflict} \\
    \multicolumn{1}{l|}{\textbf{NSERP-ODD}} & VODD & 42 &  [-0.5, 0]  & [-0.5, 1.0] & 27,865 & 123 & Standard vs Target vs Novel \\
    \multicolumn{1}{l|}{\textbf{PD-SIM}} & SIM & 147 &  [-0.3, -0.2]  & [-0.5, 1.0] & 55,921 & 60 & HC vs PD \\
    \multicolumn{1}{l|}{\textbf{PD-ODD}} & VODD & 145 & [-0.3, -0.2] & [-0.5, 1.0] & 34,464 & 60 & HC vs PD \\
    \multicolumn{1}{l|}{\textbf{ADHD-WMRI}} & N-Back, GoNogo  & 59 &  [-0.2, 0]  & [-0.2, 0.65] & 21,832 & 21 & HC vs ADHD \\
    \multicolumn{1}{l|}{\textbf{SCPD}} & SIM & 56 &  [-0.3, -0.2]  & [-0.5, 1.0] & 10,224 & 59 & HC vs PD \\
    \multicolumn{1}{l|}{\textbf{RLPD}} & RL & 56 &  [-0.2, 0]  & [-2.0, 1.0] & 14,325 & 56 & HC vs PD \\
    \multicolumn{1}{l|}{\textbf{AOPD}} & AODD & 50 &  [-0.2, 0]  & [-0.2, 0.8] & 9,830 & 59 & HC vs PD \\

    \bottomrule
    \end{tabular}
    } 
    \vspace{-5mm}
\end{table*}

\textbf{Deep Learning Methods:} Deep learning methods with diverse architectures are widely used for EEG decoding. EEGNet~\cite{lawhern2018eegnet} is a classic lightweight convolutional model, while EEG-Inception~\cite{santamaria2020eeg} extends temporal convolution with depthwise spatial blocks for ERP-based BCIs. Transformer-based approaches, such as EEG-Transformer~\cite{xie2022transformer} and EEGConformer~\cite{kasthuri2024eeg}, leverage self-attention to capture long-range temporal dependencies and global contextual information in EEG signals. Neuro-BERT~\cite{wu2022neuro} introduces masked autoencoding for robust EEG representation learning, LGGNet~\cite{ding2023lggnet} explicitly models local-global brain functional graphs, MOCNN~\cite{jin2024mocnn} proposes multiscale octave convolution for ERP classification, and EEGMamba~\cite{gui2024eegmamba} employs bidirectional Mamba modules for efficient long-sequence EEG modeling.

\textbf{Foundation Models:} Foundation models, which have achieved remarkable success in CV and NLP domains, are increasingly being explored for EEG representation learning. LaBraM~\cite{jiang2024large} introduces a large-scale EEG foundation model that employs a neural tokenizer to reconstruct spectral representations during pre-training. EEGPT~\cite{wang2024eegpt} proposes a masked self-supervised learning framework based on high signal-to-noise EEG representations, enhanced by spatio-temporal representation alignment. FORMED~\cite{huang2024repurposing} repurposes a general time series foundation model to EEG downstream tasks.  NeuroLM~\cite{jiang2024neurolm} conceptualizes EEG signals as a foreign language and adopts language-model-inspired architectures to learn unified EEG representations. LUNA~\cite{doner2025luna} presents a topology-agnostic EEG foundation model using cross-attention mechanisms, supporting robust learning under heterogeneous channel montages. CSBrain~\cite{zhou2025csbrain} explores cross-scale spatiotemporal tokenization with structured sparse attention to model dependencies between local and global brain regions.

\section{Datasets}
\label{sec:preliminary}

\subsection{Datasets Selection}

In light of the limited availability of public ERP datasets, we systematically searched all accessible resources known to the authors, including platforms such as OpenNeuro, FigShare, and PhysioNet. We identified 12 datasets, each with a sufficient subject size (40+ subjects) to ensure robust statistical evaluation, including \textbf{ADHD-WMRI}~\cite{breitling2020economical}, \textbf{AOPD}~\cite{cavanagh2021eeg}, \textbf{CESCA-AOOD}, \textbf{CESCA-VOOD}, \textbf{CESCA-FLANKER}~\cite{isbell2025cognitive}, \textbf{mTBI-ODD}~\cite{cavanagh2019erps}, \textbf{NSERP-MSIT}, \textbf{NSERP-ODD}~\cite{dzianok2022nencki}, \textbf{PD-ODD}~\cite{singh2023evoked}, \textbf{PD-SIM}~\cite{singh2023evoked}, \textbf{RLPD}~\cite{brown2020eeg}, and \textbf{SCPD}~\cite{singh2018mid}. These datasets span a broad range of ERP paradigms, such as oddball (ODD), Simon conflict (SIM), and flanker, as well as various clinical conditions, including attention deficit hyperactivity disorder (ADHD), Parkinson’s disease (PD), traumatic brain injury (TBI), and aging-related cognitive decline. To ensure fair and consistent benchmarking, all datasets are preprocessed using a unified ERP preprocessing pipeline.

Among them, 6 datasets, including CESCA-AOOD, CESCA-VOOD, CESCA-FLANKER, mTBI-ODD, NSERP-MSIT, and NSERP-ODD, are used for ERP stimulus classification, which aims to distinguish cognitive responses elicited by different experimental conditions, such as target vs non-target and congruent vs incongruent stimuli. The remaining 6 datasets, including PD-SIM, PD-ODD, RLPD, SCPD, AOPD, and ADHD-WMRI, are employed for brain disease detection tasks.

\subsection{Data Preprocessing Pipeline}
\label{sub:pipeline}

\begin{figure*}[h]
    \centering
    \includegraphics[width=1.0\linewidth]{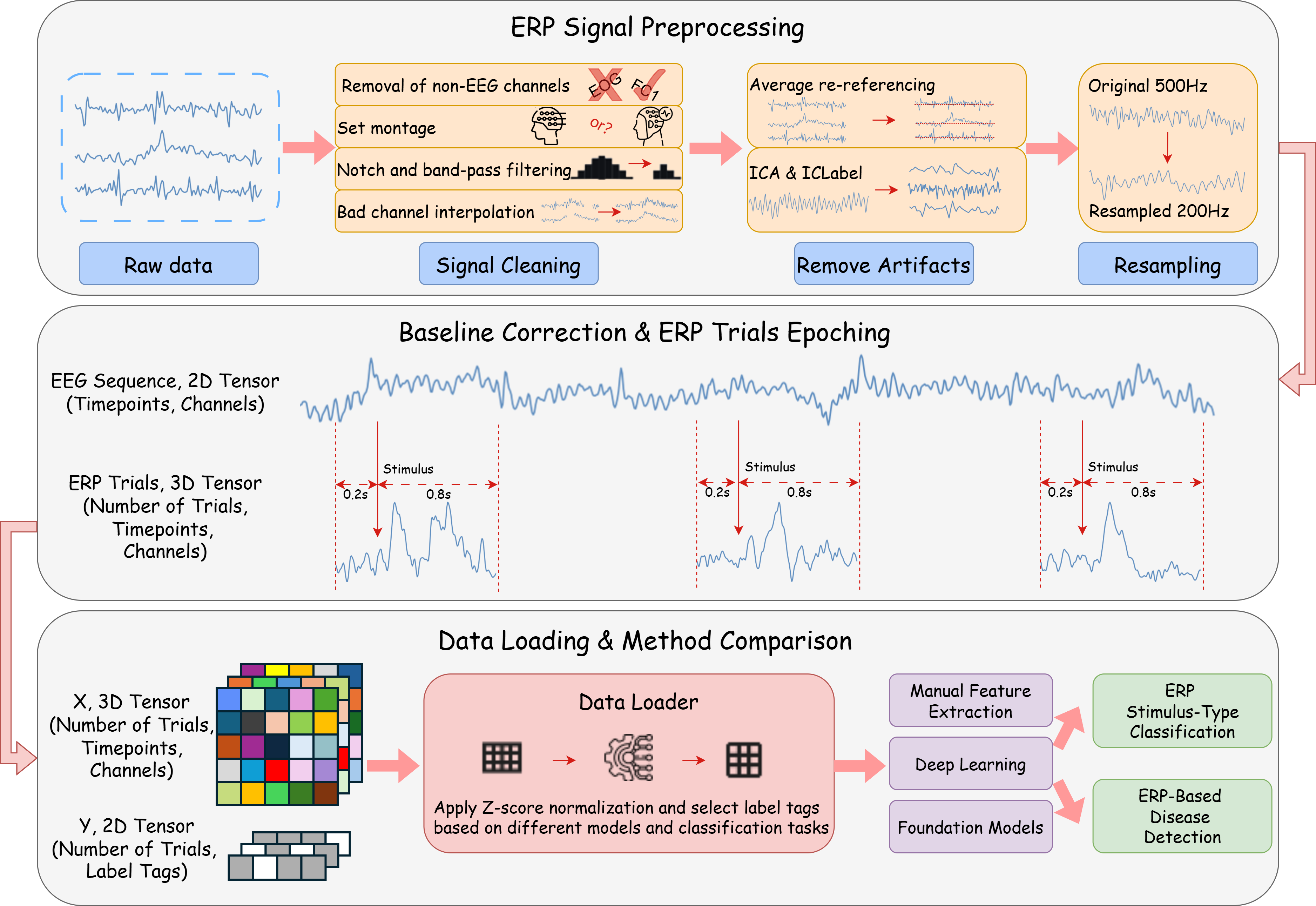}
    \caption{\textbf{Pipeline of ERP Preprocessing.} Input raw EEG data are preprocessed with a unified pipeline to get ERP trials, including removal of non-EEG channels, notch and band-pass filtering, bad channel interpolation, average re-referencing, artifact removal, resampling, baseline correction, trial epoching, and Z-score normalization. The processed ERP trials are loaded and passed to various models for training and classification, including manual feature extraction, supervised deep learning trained from scratch, and foundation models with pre-trained weights.
    }
    \label{fig:erp_pipeline}
    \vspace{-5mm}
\end{figure*}

To get ERP trials for training, we apply the following unified preprocessing pipeline (Figure~\ref{fig:erp_pipeline}).
\textbf{1) Removal of non-EEG channels:} All non-EEG channels are removed, such as EOG or coordinate information. \textbf{2) Notch and band-pass filtering:} A notch filter at 50 Hz or 60 Hz is applied to suppress line noise, followed by a band-pass filter between 0.5 Hz and 45 Hz. \textbf{3) Bad channel interpolation:} Bad channels are interpolated when marked. \textbf{4) Average re-referencing:} Average re-referencing is applied to reduce global noise and potential baseline shifts. \textbf{5) Artifact removal:} Independent component analysis (ICA), combined with ICLabel~\cite{pion2019iclabel}, is used to automatically identify and remove components associated with eye blinks, muscle activity, and cardiac artifacts. \textbf{6) Resampling:} All recordings are resampled to a uniform sampling rate of 200 Hz. \textbf{7) Baseline correction:} Baseline correction is applied to enhance stimulus-related ERP components relative to background noise. \textbf{8) Trial epoching:} EEG recordings are then segmented into ERP trials based on stimulus events, and the non-ERP recordings are discarded (e.g, resting state). The baseline correction and epoch window length vary across ERP tasks and datasets. \textbf{9) Z-Score Normalization:} Each channel of every segmented ERP trial is normalized independently to zero mean and unit variance. The statistics of the processed datasets are summarized in Table~\ref{tab:processed_data}. More details of per-dataset preprocessing are provided in our GitHub repository.

\section{Method}
\label{sec:method}

\begin{table*}[t]
    \centering
    \renewcommand{\arraystretch}{1.20}
    \caption{\textbf{Summary of Methods.} We report the total number of parameters (which may vary by dataset), the backbone architecture, whether pre-training is used, and the key methodological innovations. GNN stands for graph neural networks. }
    \vspace{-2mm}
    \label{tab:method_summary}

    \small
    \begin{tabularx}{\textwidth}{@{}l|c|c|c|X@{}}
    \toprule
    
    \textbf{Methods} 
    & \makecell{\textbf{Parameters} \\ \textit{(Million)}} 
    & \textbf{Backbone}
    & \textbf{Pre-training}
    & \multicolumn{1}{c}{\textbf{Method Innovations}} \\

    \midrule
    
    \multirow{1}{*}{\textbf{TCN}} 
    & \multirow{1}{*}{1.1}
    & \multirow{1}{*}{CNN}
    & \multirow{1}{*}{No}
    & Uses 1D causal dilated convolutions to enlarge the receptive field. \\
    \cline{1-5}

    \multirow{2}{*}{\textbf{ModernTCN}} 
    & \multirow{2}{*}{1.2 -- 13}
    & \multirow{2}{*}{CNN}
    & \multirow{2}{*}{No}
    & Combines depthwise-pointwise convolutions with channel-independent embeddings to capture long-range and cross-variable dependencies. \\
    \cline{1-5}

    \multirow{2}{*}{\textbf{TimesNet}} 
    & \multirow{2}{*}{2.3}
    & \multirow{2}{*}{CNN}
    & \multirow{2}{*}{No}
    & Transforms 1D time-series into multi-periodicity 2D representations and applies 2D-CNN to capture both intra- and inter-period temporal variations. \\
    \cline{1-5}

    \multirow{2}{*}{\textbf{PatchTST}} 
    & \multirow{2}{*}{0.8 -- 2.5}
    & \multirow{2}{*}{Transformer}
    & \multirow{2}{*}{No}
    & Introduces single-channel patch embedding and channel-independent modeling for time-series Transformers. \\
    \cline{1-5}

    \multirow{2}{*}{\textbf{iTransformer}} 
    & \multirow{2}{*}{0.8 -- 0.9}
    & \multirow{2}{*}{Transformer}
    & \multirow{2}{*}{No}
    & Treats each variable as an independent token, enabling self-attention to model inter-variable dependencies in multivariate time series. \\
    \cline{1-5}

    \multirow{2}{*}{\textbf{Medformer}} 
    & \multirow{2}{*}{2.4 -- 4.7}
    & \multirow{2}{*}{Transformer}
    & \multirow{2}{*}{No}
    & A Transformer model for medical time-series that uses cross-channel patching, multi-granularity embeddings, and hierarchical self-attention. \\
    \cline{1-5}

    \multirow{2}{*}{\textbf{MedGNN}} 
    & \multirow{2}{*}{4.2 -- 4.7}
    & \multirow{2}{*}{\makecell{GNN \& \\ Transformer }}
    & \multirow{2}{*}{No}
    & A multi-scale graph model that uses adaptive graphs and frequency-domain convolution and graph transformers for medical time-series learning. \\
    \cline{1-5}

    \multirow{2}{*}{\textbf{EEGNet}} 
    & \multirow{2}{*}{0.002}
    & \multirow{2}{*}{CNN}
    & \multirow{2}{*}{No}
    & A compact CNN architecture that uses temporal, depthwise, and separable convolutions to efficiently capture temporal and spatial features in EEG. \\
    \cline{1-5}

    \multirow{2}{*}{\textbf{EEGInception}} 
    & \multirow{2}{*}{0.8}
    & \multirow{2}{*}{CNN}
    & \multirow{2}{*}{No}
    & Uses Inception-style multi-scale temporal convolutions with separable convolutions to capture EEG features across multiple temporal scales. \\
    \cline{1-5}

    \multirow{2}{*}{\textbf{EEGConformer}} 
    & \multirow{2}{*}{1.1 -- 2.9}
    & \multirow{2}{*}{\makecell{CNN \& \\ Transformer }}
    & \multirow{2}{*}{No}
    & Combines convolutional spatial-temporal embedding with Transformer self-attention to capture local EEG patterns and global dependencies. \\
    \cline{1-5}

    \multirow{3}{*}{\textbf{BIOT}} 
    & \multirow{3}{*}{3.2}
    & \multirow{3}{*}{Transformer}
    & \multirow{3}{*}{Yes}
    & A foundation model for biomedical signals that single-channel patch embeddings with channel tags and self-attention to capture temporal and spatial dependencies. Pre-trained on 2 EEG datasets and 1 ECG dataset.\\
    \cline{1-5}

    \multirow{3}{*}{\textbf{LaBraM}} 
    & \multirow{3}{*}{5.8}
    & \multirow{3}{*}{Transformer}
    & \multirow{3}{*}{Yes}
    & An EEG foundation model that learns EEG representations via vector-quantized neural tokenization and masked EEG reconstruction with a Transformer. Pre-trained on 16 datasets with 2500+ hours of EEG. \\
    \cline{1-5}

    \multirow{3}{*}{\textbf{CBraMod}} 
    & \multirow{3}{*}{8.1 -- 14}
    & \multirow{3}{*}{Transformer}
    & \multirow{3}{*}{Yes}
    & An EEG foundation model using a criss-cross Transformer with parallel spatial and temporal attention to model channel and temporal dependencies. Pre-trained on the TUEG dataset with 9000+ hours of EEG. \\

    \bottomrule
    \end{tabularx}

    \vspace{-5mm}
\end{table*}

\subsection{Manual Extract Features}
Widely used features in spontaneous EEG and ERP data analysis. We manually extract these features and apply a linear projection layer to them for ERP task classification.

\subsubsection{EEG Feature Extraction}

We compute \textbf{31 widely used features}, including 10 time-domain statistical features, 11 frequency band features, 7 spectral distribution features, and 3 data complexity features. 

\textbf{Time-Domain Statistical Features:}
These include the mean, median, minimum, maximum, skewness, kurtosis, root mean square (RMS), interquartile range, and standard deviation~\cite{tzimourta2019eeg, tzimourta2019analysis, kulkarni2017extracting, kanda2014clinician}. Statistical features are designed to capture fundamental time-domain statistical characteristics of EEG signals.

\textbf{Frequency-Band Power Features:}
This category consists of absolute power in the $\delta$, $\theta$, $\alpha$, and $\beta$ bands, total power, the $\theta/\alpha$ ratio, the $\alpha/\beta$ ratio, and the relative power of the four frequency bands~\cite{schmidt2013index, liu2016multiple, kanda2014clinician}. Absolute band power is computed by numerically integrating the power spectral density (PSD) within each frequency band, providing a direct measure of oscillatory energy associated with specific neuro processes.

\textbf{Spectral Distribution Features:}
The spectral features include spectral centroid, spectral roll-off frequency, peak frequency, peak power, mean frequency, median frequency, and spectral flatness~\cite{wang2017enhanced, cassani2014effects, wang2015multiple}. These features characterize the overall distribution and structure of spectral energy. All spectral features are estimated using Welch’s method for PSD estimation, followed by numerical integration and computations.

\textbf{Complexity features:}
This category includes normalized \& non-normalized Shannon spectral entropy and Tsallis entropy~\cite{garn2015quantitative, azami2019multiscale, tylova2018unbiased}. These features are derived from the normalized PSD, which is interpreted as a probability distribution. Shannon spectral entropy quantifies spectral disorder and complexity, with higher values indicating more uniformly distributed spectral energy, while Tsallis entropy generalizes Shannon entropy by incorporating non-extensive properties, providing greater sensitivity to nonlinear and non-stationary signal characteristics.

\subsubsection{ERP Feature Extraction}
We compute a total of \textbf{91 features}, including 75 temporal pyramid pooling features, 4 peak-related Features, 9 frequency-band \& complexity features, and 3 Hjorth parameters. 

\textbf{Temporal Pyramid Pooling:} five statistical measures, including mean, standard deviation, RMS, line length, and peak-to-peak amplitude~\cite{waser2013eeg, tylova2013predictive, mora2019scale, barry2000eeg}, are computed across multiple temporal scales, yielding $5 \times \sum \text{levels}$ feature dimensions. We set the number of levels equal to 15. Temporal Pyramid Pooling is employed to capture ERP dynamics at multiple temporal resolutions. This approach jointly captures global waveform structure and localized transients, improving robustness to temporal variability and latency jitter.

\textbf{Peak-Related Features:} amplitudes of positive and negative peaks and their normalized latencies~\cite{mak2012eeg, li2013identifying, wallace2017eeg}. Peak-related features are extracted to characterize salient ERP components. The maximum positive and minimum negative peak amplitudes reflect the strength of event-related neural responses. The total analysis window-normalized latencies to account for ERP duration variability and ensure comparability across trials.

\textbf{Frequency \& Complexity Features:} relative power in the $\delta$, $\theta$, $\alpha$, and $\beta$ bands, total power, spectral centroid, spectral flatness, median frequency, and normalized spectral entropy~\cite{fraga2013characterizing, tait2019network, waser2016quantifying, trambaiolli2011improving}. Although ERPs are time-locked signals, frequency-domain analysis provides complementary information about oscillatory patterns elicited by stimuli. Relative band power and total power summarize energy distribution across canonical frequency bands, while spectral centroid, spectral flatness, and median frequency describe the overall frequency structure of the ERP signal. Normalized spectral entropy is computed to quantify the complexity and dispersion of ERP spectral content. 

\textbf{Hjorth parameters:} Hjorth parameters includes activity, mobility, and complexity~\cite{peng2021early, mehmood2022eeg, deng2025diagnosis}. These features serve as compact descriptors of ERP signal characteristics. Activity reflects the overall signal amplitude by measuring signal variance. Mobility characterizes the mean frequency content by relating the variance of the first derivative to the variance of the signal itself. Complexity quantifies waveform shape variation relative to a sinusoidal signal, thereby providing insight into the structural intricacy of ERP dynamics.

\subsection{Deep Learning Models}
Deep learning methods for general time-series, medical time-series, and EEG data classification. They are trained from scratch in a fully supervised learning manner. We select classic or state-of-the-art methods with publicly available code.

\subsubsection{TCN} 
TCN~\cite{lea2017temporal} is a convolutional architecture designed explicitly for time-series modeling. They extend residual networks by incorporating causal and dilated convolutions, enabling effective learning of long-range temporal dependencies. By flexibly controlling receptive field size, TCNs balance model capacity and computational efficiency, thereby making them adaptable to time-series data across diverse domains. TCNs remain a strong and widely adopted baseline for time-series classification tasks, including EEG analysis.

\subsubsection{ModernTCN}
ModernTCN~\cite{luo2024moderntcn} is a convolutional architecture designed for general time-series analysis and demonstrates strong performance, particularly in classification. It combines depthwise convolutions with multiple pointwise convolutions and introduces channel-independent embedding mechanisms. This design effectively addresses key limitations of convolutional models in multi-variate time-series tasks, where performance is often constrained by small receptive fields and difficulties in modeling inter-variable dependencies.

\subsubsection{TimesNet}
TimesNet~\cite{wu2022timesnet} is a convolutional model for time-series analysis, especially on classification tasks. It transforms one-dimensional time-series data into two-dimensional tensors across multiple periodicities (where rows and columns represent intra- and inter-period variations, respectively), thereby leveraging the strengths of 2D convolutional networks for time-series analysis. The model's core component, TimesBlock, employs a parameter-efficient Inception module to extract complex temporal variations from 2D tensors.

\subsubsection{PatchTST}
PatchTST~\cite{nie2022time} is a transformer-based model for time series prediction and representation learning, with its core innovation lying in a dual design of patching and channel independence. The model decomposes multi-channel data into multiple one-channel patches, treating them as semantically informative sub-sequence fragments for input. This approach preserves local features while significantly reducing computational complexity.

\subsubsection{iTransformer} 
The iTransformer~\cite{liu2023itransformer} proposes a novel data-embedding method based on the transformer architecture for multi-variate time-series analysis. Its distinctive feature is the reversal of the conventional tokenization strategy. Instead of using multi-channel samples at a single time point into a single temporal token, iTransformer treats each channel/variate as an independent variable token. This adjustment enables the self-attention module to model correlations between variables directly, thereby enhancing interpretability.

\subsubsection{Medformer} 
Medformer~\cite{wang2024medformer} is a multi-granularity patching Transformer model designed explicitly for medical time-series data (e.g., EEG, ECG). It captures correlations across signal channels via cross-channel patching and extracts features at multiple temporal scales via multi-granularity embedding. Subsequently, it applies two-stage (intra-granularity \& inter-granularity) multi-granularity self-attention to learn features. Medformer is a modified Transformer model adapted for medical electro-biological time-series data.

\subsubsection{MedGNN} 
MedGNN~\cite{fan2025towards} proposes a multi-scale spatio-temporal graph learning framework for medical time series such as electroencephalograms. This framework models cross-channel spatial relationships by constructing adaptive graph structures and integrating multi-scale temporal features via frequency-domain convolution. MedGNN further employs a differential-based attention mechanism and multi-scale graph transformers to capture local temporal dynamics and global graph dependencies in a synergistic manner.

\subsubsection{EEGNet} 
As a classic deep learning application in the EEG domain, EEGNet~\cite{lawhern2018eegnet} efficiently extracts temporal and spatial features via convolutional operations. By stacking 2D temporal, depthwise, and separable convolution modules, it captures frequency features and feature relationships across different electrode channels. Its structure is compact and requires low space and computational cost.

\subsubsection{EEGInception}
EEGInception~\cite{santamaria2020eeg} pioneers the integration of Inception modules~\cite{chollet2017xception} into deep learning models for ERP detection tasks. By employing multi-scale temporal convolutions to extract features across different temporal scales within EEG signals concurrently, and incorporating separable convolutions, batch normalization, and Dropout, it constructs a lightweight and efficient network architecture.

\subsubsection{EEGConformer} 
EEGConformer~\cite{song2022eeg} is a Transformer-based architecture that integrates convolutional networks for spatio-temporal feature embedding with self-attention mechanisms for EEG representation learning. This design combines the strengths of convolutional networks in capturing local spatio-temporal patterns and self-attention in modeling global dependencies, while maintaining parameter efficiency and enhancing the ability to represent complex EEG structures.

\subsection{Foundation Models}
Medical time-series and EEG foundation models. We use their pre-trained weights for fine-tuning on our ERP tasks. We select state-of-the-art methods with publicly available code.

\subsubsection{BIOT}~\cite{yang2023biot} 
BIOT introduces a large foundation model for the flexible processing of multi-channel biomedical signals of varying lengths and channels. It adopts language-model-style segmentation to convert single-channel signals into patch embeddings, augments them with channel tags, and applies self-attention to capture temporal and spatial dependencies. This design supports multi-stage training and remains effective under missing-channel or missing-segment scenarios. We load their pre-trained weights for fine-tuning.

\subsubsection{LaBraM} 
LaBraM~\cite{jiang2024large} is a foundation model for learning universal EEG representations through large-scale unsupervised pre-training. The framework's core lies in a vector-quantization neural spectral prediction method: First, a neural tokenizer is trained to encode continuous EEG segments into a compact neural lexicon by reconstructing the Fourier spectrum of the raw EEG signals. Subsequently, a neural Transformer is pre-trained using masked EEG modeling (randomly masking portions of EEG segments and predicting their corresponding neural lexical entries) based on this lexicon. Equipped with learnable temporal and spatial embeddings, this model adaptively learns from EEG inputs of arbitrary channel counts and durations. This method uses more than \textbf{2,500 hours} of EEG data for pre-training. We load their pre-trained weights for fine-tuning.

\begin{table*}[t]
    \centering
    \scriptsize 
    \caption{\textbf{ERP Stimulus Classification Results.} Classifying ERP stimulus types such as target, non-target, and distractor in Oddball, or congruent and incongruent in Flanker. \textcolor{myred}{\textbf{Top-1}}, \textcolor{myblue}{\underline{Top-2}}, and \textcolor{mygreen}{Top-3} results are highlighted in red, blue, and green.
    }
    \vspace{-2mm}
    \label{tab:erp_stimulus_results}
    \resizebox{\textwidth}{!}{
    \begin{tabular}{@{}ll|ccc|ccc|ccc@{}}
    \toprule

    \multicolumn{2}{l|}{\textbf{Datasets}}
    & \multicolumn{3}{c|}{\makecell{\textbf{CESCA-AODD} \\ \textit{(38,151 Trials)} \\ \textit{(127 Subjects, 2 Classes)}} }
    & \multicolumn{3}{c|}{\makecell{\textbf{CESCA-VODD} \\ \textit{(20,419 Trials)} \\ \textit{(127 Subjects, 2 Classes)}} }
    & \multicolumn{3}{c}{\makecell{\textbf{CESCA-FLANKER} \\ \textit{(29,774 Trials)} \\ \textit{(73 Subjects, 2 Classes)}} }
    \\ \midrule

    \multicolumn{2}{l|}{\diagbox{\textbf{Methods}}{\textbf{Metrics}}} & \textbf{Acccuracy} & \textbf{F1 Score} & \textbf{AUROC} & \textbf{Acccuracy} & \textbf{F1 Score} & \textbf{AUROC} & \textbf{Acccuracy} & \textbf{F1 Score} & \textbf{AUROC} \\ \midrule

    \multicolumn{2}{l|}{\textbf{EEG Features}}  &  \textcolor{myblue}{\underline{77.06\std{0.96}}} & 46.44\std{1.03}  & 50.71\std{0.38}  & 77.26\std{1.02} & 50.31\std{0.84}   & 56.73\std{1.16} & 55.38\std{1.12} & 55.13\std{1.12}  & 57.74\std{1.84}  \\
    \multicolumn{2}{l|}{\textbf{ERP Features}}  & 75.43\std{3.46} & 47.74\std{1.66}  & 52.35\std{1.67}  & \textcolor{mygreen}{81.71\std{0.43}} & 63.68\std{1.10}   & 76.67\std{1.83} & \textcolor{myblue}{\underline{64.12\std{1.31}}} & 63.98\std{1.23}  & 69.66\std{1.57}  \\

    \midrule
    \multicolumn{2}{l|}{\textbf{TCN}}  & 75.07\std{0.40} & 50.74\std{0.99}  & 56.39\std{0.95}  & \textcolor{myred}{\textbf{82.03\std{0.69}}} & 67.00\std{1.12}   & \textcolor{mygreen}{78.13\std{1.42}} & 63.16\std{1.11} & 62.89\std{1.30}  & 68.62\std{1.43}  \\
    \multicolumn{2}{l|}{\textbf{ModernTCN}}  & 75.20\std{0.69} & 53.93\std{0.82}  & 60.36\std{0.98}  & \textcolor{myblue}{\underline{81.91\std{0.51}}} & 66.85\std{3.11}   & \textcolor{myblue}{\underline{78.18\std{1.01}}} & 64.07\std{0.58} & \textcolor{mygreen}{64.01\std{0.59}}  & \textcolor{mygreen}{69.90\std{1.39}}  \\
    \multicolumn{2}{l|}{\textbf{TimesNet}}  & 75.17\std{0.58} & 54.38\std{1.34}  & 60.91\std{0.74}  & 81.13\std{1.53} & \textcolor{mygreen}{67.65\std{1.55}}   & 77.62\std{1.42} & 63.20\std{0.64} & 63.09\std{0.59}  & 68.29\std{1.25}  \\
    \multicolumn{2}{l|}{\textbf{PatchTST}}  &  \textcolor{mygreen}{76.41\std{1.41}} & 53.65\std{1.70}  &  \textcolor{myblue}{\underline{61.98\std{3.11}}}  & 80.97\std{1.13} & \textcolor{myblue}{\underline{68.19\std{1.72}}}   & 78.13\std{1.02} & 63.81\std{0.77} & 63.64\std{0.81}  & 69.61\std{1.33}  \\
    \multicolumn{2}{l|}{\textbf{iTransformer}}  & 74.98\std{1.22} & 53.58\std{0.63}  & 60.30\std{1.21}  & 81.39\std{0.49} & 66.20\std{1.12}   & 76.37\std{0.86} & 63.77\std{1.22} & 63.69\std{1.19}  & 69.35\std{1.68}  \\
    \multicolumn{2}{l|}{\textbf{Medformer}}  & 74.41\std{0.90} &  \textcolor{myred}{\textbf{57.18\std{0.42}}}  &  \textcolor{myred}{\textbf{62.87\std{1.06}}}  & 80.25\std{1.02} & 66.58\std{0.88}   & 75.85\std{1.23} & 63.56\std{0.98} & 63.42\std{1.06}  & 69.26\std{1.41}  \\
    \multicolumn{2}{l|}{\textbf{MedGNN}}  & 74.68\std{0.38} & \textcolor{mygreen}{55.29\std{0.77}}  & 60.92\std{0.68}  & 81.55\std{0.86} & 67.52\std{1.03}   & 77.95\std{1.16} & \textcolor{myred}{\textbf{64.43\std{1.21}}} & \textcolor{myred}{\textbf{64.33\std{1.22}}}  & \textcolor{myred}{\textbf{70.45\std{1.49}}}  \\
    \multicolumn{2}{l|}{\textbf{EEGNet}}  & \textcolor{myred}{\textbf{79.12\std{0.02}}} & 44.17\std{0.01}  & 50.74\std{0.37}  & 77.83\std{2.28} & 47.95\std{3.75}   & 57.28\std{1.72} & 51.86\std{0.62} & 51.15\std{1.25}  & 52.89\std{0.78}  \\
    \multicolumn{2}{l|}{\textbf{EEGInception}}  & 72.16\std{8.83} & 46.59\std{2.31}  & 51.73\std{0.85}  & 77.36\std{1.76} & 62.58\std{1.31}   & 71.14\std{2.17} & 59.07\std{0.50} & 57.99\std{1.56}  & 63.65\std{0.89}  \\
    \multicolumn{2}{l|}{\textbf{EEGConformer}}  & 74.74\std{0.65} & 54.40\std{0.46}  & 61.01\std{1.45}  & 81.31\std{0.67} & \textcolor{myred}{\textbf{69.64\std{1.35}}}   & \textcolor{myred}{\textbf{79.58\std{1.21}}} & \textcolor{mygreen}{64.09\std{1.18}} & \textcolor{myblue}{\underline{64.01\std{1.18}}}  & 69.83\std{1.58}  \\

    \midrule
    \multicolumn{2}{l|}{\textbf{BIOT}}  & 73.89\std{2.75} & 48.40\std{2.20}  & 49.84\std{0.58}  & 73.94\std{1.64} & 54.28\std{1.41}   & 58.63\std{1.24} & 54.51\std{1.15} & 54.20\std{1.09}  & 56.10\std{1.31}  \\
    \multicolumn{2}{l|}{\textbf{LaBraM}}  & 74.33\std{1.02} &  \textcolor{myblue}{\underline{55.90\std{1.13}}}  &  \textcolor{mygreen}{61.95\std{1.26}}  & 80.21\std{0.51} & 65.75\std{0.97}   & 75.31\std{1.10} & 63.40\std{1.33} & 63.27\std{1.31}  & 69.02\std{1.65}  \\
    \multicolumn{2}{l|}{\textbf{CBraMod}}  & 76.09\std{0.61} & 53.37\std{0.63}  & 59.76\std{0.55}  & 80.85\std{0.82} & 66.25\std{1.25}   & 76.00\std{1.37} & 64.01\std{1.37} & 63.87\std{1.48}  & \textcolor{myblue}{\underline{69.97\std{1.40}}}  \\
    \midrule

    \midrule
    \multicolumn{2}{l|}{\textbf{Datasets}}
    & \multicolumn{3}{c|}{\makecell{\textbf{mTBI-ODD} \\ \textit{(24,885 Trials)} \\ \textit{(96 Subjects, 3 Classes)}} }
    & \multicolumn{3}{c|}{\makecell{\textbf{NSERP-MSIT} \\ \textit{(16,729 Trials)} \\ \textit{(42 Subjects, 4 Classes)}} }
    & \multicolumn{3}{c}{\makecell{\textbf{NSERP-ODD}  \\ \textit{(27,865 Trials)} \\ \textit{(42 Subjects, 3 Classes)}} }
    \\ \midrule

    \multicolumn{2}{l|}{\diagbox{\textbf{Methods}}{\textbf{Metrics}}} & \textbf{Acccuracy} & \textbf{F1 Score} & \textbf{AUROC} & \textbf{Acccuracy} & \textbf{F1 Score} & \textbf{AUROC} & \textbf{Acccuracy} & \textbf{F1 Score} & \textbf{AUROC} \\ \midrule

    \multicolumn{2}{l|}{\textbf{EEG Features}}  & 68.11\std{2.22} & 35.10\std{2.39}  & 60.61\std{4.07}  & 26.87\std{0.94} & 26.42\std{0.93}   & 52.58\std{1.32} & 72.60\std{2.11} & 36.18\std{2.17}  & 64.93\std{3.44}  \\
    \multicolumn{2}{l|}{\textbf{ERP Features}}  & 77.34\std{0.79} & 58.75\std{2.23}  & 81.65\std{0.95}  & 36.09\std{2.13} & 35.42\std{2.03}   & 63.69\std{2.32} & 82.26\std{1.63} & 62.35\std{3.00}  & 87.12\std{2.16}  \\

    \midrule
    \multicolumn{2}{l|}{\textbf{TCN}}  & 77.87\std{1.15} & 62.44\std{1.77}  & 83.74\std{1.32}  & 35.26\std{1.55} & 34.45\std{1.76}   & 63.37\std{1.95} & \textcolor{mygreen}{84.17\std{1.25}} & \textcolor{mygreen}{66.54\std{2.54}}  & \textcolor{mygreen}{90.17\std{1.68}}  \\
    \multicolumn{2}{l|}{\textbf{ModernTCN}}  & \textcolor{myred}{\textbf{79.16\std{0.98}}} & 63.77\std{1.33}  & \textcolor{myblue}{\underline{84.80\std{1.18}}}  & 36.94\std{2.38} & 36.72\std{2.37}   & 65.04\std{2.63} & 83.28\std{0.97} & 63.21\std{2.68}  & 88.82\std{1.32}  \\
    \multicolumn{2}{l|}{\textbf{TimesNet}}  & 78.09\std{1.28} & 63.66\std{2.68}  & 84.30\std{1.35}  & 36.86\std{2.40} & 36.44\std{2.24}   & 65.31\std{2.47} & 82.85\std{0.94} & 63.54\std{2.03}  & 88.74\std{1.30}  \\
    \multicolumn{2}{l|}{\textbf{PatchTST}}  & 78.25\std{1.64} & 63.57\std{1.91}  & 84.49\std{1.25}  & 36.59\std{2.58} & 36.22\std{2.67}   & 64.13\std{2.87} & 82.94\std{1.01} & 63.70\std{1.89}  & 88.34\std{1.66}  \\
    \multicolumn{2}{l|}{\textbf{iTransformer}}  & 78.42\std{1.13} & \textcolor{myblue}{\underline{64.36\std{1.48}}}  & 84.54\std{1.40}  & 36.40\std{2.40} & 36.03\std{2.34}  & 64.68\std{2.51} & 81.74\std{1.36} & 61.85\std{2.43}  & 87.52\std{1.69}  \\
    \multicolumn{2}{l|}{\textbf{Medformer}}  & 77.82\std{1.33} & \textcolor{mygreen}{63.99\std{1.37}}  & 83.93\std{1.20}  & 37.64\std{2.38} & \textcolor{mygreen}{37.29\std{2.45}}   & 65.69\std{2.81} & 81.38\std{0.70} & 63.33\std{1.77}  & 87.64\std{1.15}  \\
    \multicolumn{2}{l|}{\textbf{MedGNN}}  & \textcolor{mygreen}{78.51\std{0.97}} & 63.92\std{1.64}  & \textcolor{mygreen}{84.74\std{1.05}}  & \textcolor{mygreen}{37.74\std{2.26}} & 37.20\std{2.15}   & \textcolor{myblue}{\underline{66.40\std{2.55}}} & 83.66\std{1.19} & 65.13\std{2.77}  & 90.08\std{1.39}  \\
    \multicolumn{2}{l|}{\textbf{EEGNet}}  & 63.86\std{4.92} & 34.00\std{3.20}  & 58.90\std{1.39}  & 25.83\std{1.28} & 20.92\std{1.98}   & 50.24\std{0.87} & 60.16\std{2.62} & 35.35\std{1.35}  & 60.42\std{0.94}  \\
    \multicolumn{2}{l|}{\textbf{EEGInception}}  & 65.07\std{4.86} & 48.70\std{1.08}  & 70.36\std{0.89}  & 31.43\std{2.09} & 27.67\std{4.29}   & 58.99\std{2.58} & 75.48\std{3.77} & 58.19\std{4.68}  & 82.23\std{3.26}  \\
    \multicolumn{2}{l|}{\textbf{EEGConformer}}  & \textcolor{myblue}{\underline{78.73\std{1.70}}} & \textcolor{myred}{\textbf{65.79\std{2.15}}}  & \textcolor{myred}{\textbf{85.63\std{1.45}}}  & \textcolor{myblue}{\underline{38.59\std{2.01}}} & \textcolor{myblue}{\underline{38.39\std{2.06}}}   & \textcolor{mygreen}{66.22\std{2.25}} & \textcolor{myblue}{\underline{84.27\std{1.32}}} & \textcolor{myred}{\textbf{68.40\std{2.25}}}  & \textcolor{myred}{\textbf{90.53\std{1.43}}}  \\

    \midrule
    \multicolumn{2}{l|}{\textbf{BIOT}}  & 63.79\std{2.77} & 39.90\std{0.75}  & 60.16\std{0.63}  & 30.40\std{0.94} & 30.06\std{1.11}   & 57.04\std{1.36} & 75.50\std{1.12} & 49.83\std{1.56}  & 74.91\std{2.58}  \\
    \multicolumn{2}{l|}{\textbf{LaBraM}}  & 76.70\std{1.76} & 61.93\std{2.34}  & 83.12\std{1.44}  & 35.91\std{2.86} & 35.54\std{2.75}   & 65.09\std{3.14} & 82.15\std{1.19} & 63.51\std{2.31}  & 88.48\std{1.78}  \\
    \multicolumn{2}{l|}{\textbf{CBraMod}}  & 78.21\std{1.07} & 63.50\std{1.63}  & 84.64\std{1.27}  & \textcolor{myred}{\textbf{39.24\std{2.77}}} & \textcolor{myred}{\textbf{38.51\std{2.62}}}   & \textcolor{myred}{\textbf{67.76\std{2.83}}} & \textcolor{myred}{\textbf{84.49\std{1.24}}} & \textcolor{myblue}{\underline{67.42\std{2.98}}}  & \textcolor{myblue}{\underline{90.40\std{1.61}}}  \\

    \bottomrule
    \end{tabular}
    }
\vspace{-2mm}
\end{table*}

\subsubsection{CBraMod} 
CBraMod~\cite{wang2024cbramod} is a foundation model for EEG that adopts a criss-cross Transformer as its backbone. By employing parallel spatial and temporal attention mechanisms, it separately models dependencies between channels at the same time point and between time segments within the same channel, thereby better aligning with the inherent structure of EEG signals. Additionally, the model uses Asymmetric Conditional Position Embeddings (ACPE) to dynamically generate position embeddings, enabling it to adapt flexibly to EEG data with varying channel configurations and temporal durations. This method uses more than \textbf{9,000 hours} of EEG data for pre-training. We load their pre-trained weights for fine-tuning.

\begin{figure}[h]
    \centering
    \includegraphics[width=1.0\columnwidth]{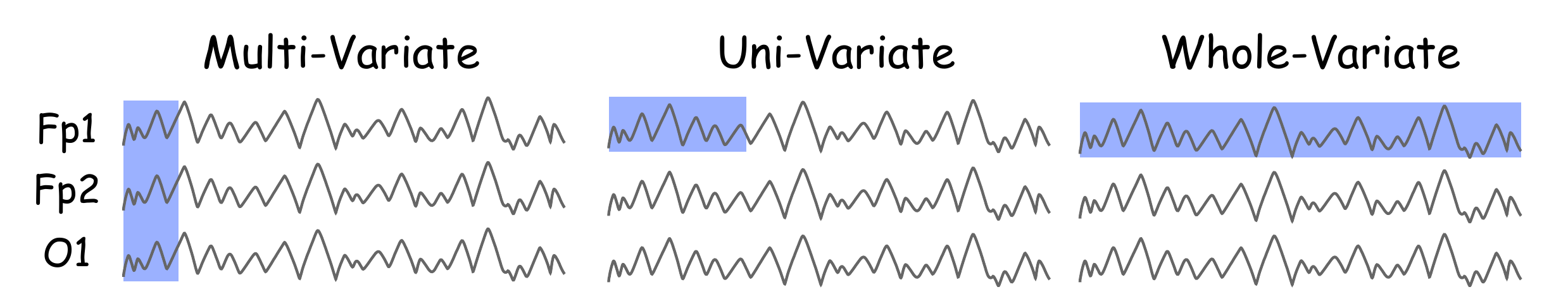}
    \caption{\textbf{Transformer Embedding Comparison.} Three commonly used EEG Transformer token embedding methods.
    }
    \label{fig:token_embedding}
    \vspace{-5mm}
\end{figure}

\subsection{Token Embedding Comparison for ERP Transformer}
Existing EEG Transformer models commonly employ three types of token embeddings: \textbf{multi-variate}, \textbf{uni-variate}, and \textbf{whole-variate}. For example, Medformer~\cite{wang2024medformer} employs multi-variate patch embedding, PatchTST~\cite{nie2022time} adopts uni-variate patch embedding, and iTransformer~\cite{liu2023itransformer} uses whole-variate patch embedding. Given a multi-variate EEG input sample $(\bm{x} \in \mathbb{R}^{T \times C}$ and a patch length $L$, multi-variate embedding uses patches $\bm{p} \in \mathbb{R}^{L \times C}$, uni-variate embedding uses patches $\bm{p} \in \mathbb{R}^{L \times 1}$, and whole-variate embedding uses patches $\bm{p} \in \mathbb{R}^{1 \times C}$. These patches are then linearly projected into token embeddings that serve as inputs to the Transformer encoder. Figure~\ref{fig:token_embedding} illustrates the three embedding strategies.

\section{Experiments}
\label{sec:experiments}

\subsection{Experimental Setups}
Batch sizes are 128 for all methods. The training epochs are set to 200 with early stopping (patience=15) based on the best F1 Score. We use AdamW with learning rates of $1\!\times\!10^{-4}$, scheduled by CosineAnnealingLR. Model performance is evaluated using Accuracy, F1 Score, and AUROC. We adopt Monte Carlo cross-validation under the \textbf{subject-independent} evaluation protocol~\cite{wang2025lead}, with a 6:2:2 ratio of subjects for training, validation, and test split. This setup ensures that no subject overlaps across splits while maintaining random subject partitioning across different seeds. For foundation model methods, including BIOT, LaBraM, and CBraMod, we fine-tune their publicly released checkpoints. Each experiment is repeated with five random seeds (41-45), and results are reported with the mean and standard deviation. All experiments are conducted on 4 NVIDIA RTX A5000 GPUs using Python 3.10 and PyTorch 2.5.1+cu121. Additional implementation details and training scripts for each method are provided in our GitHub repository.

\subsection{ERP Stimulus Classification Results}
\label{sub:erp_stimulus_results}

We first investigate stimulus-type classification in ERP data, with results summarized in Tables~\ref{tab:erp_stimulus_results}. All experiments are conducted under a subject-independent evaluation setup, aiming to identify stimulus-related ERP patterns that generalize across subjects. The \textcolor{myred}{\textbf{Top-1}}, \textcolor{myblue}{\underline{Top-2}}, and \textcolor{mygreen}{Top-3} performing methods are highlighted in red, blue, and green, respectively. Across the six datasets (CESCA-AOOD, CESCA-VOOD, CESCA-FLANKER, mTBI-ODD, NSERP-MSIT, and NSERP-ODD), the best method achieved F1 scores of 57.18\%, 69.64\%, 64.33\%, 65.79\%, 38.51\%, and 68.40\%, respectively. These results are substantially higher than random performance, indicating that meaningful and consistent stimulus-related ERP patterns can be extracted even under cross-subject evaluation. Among the three CESCA datasets, the visual oddball and flanker paradigms exhibit higher discriminability than the auditory oddball paradigm. Since these datasets share the same group of subjects and recording conditions, this observation suggests that visual oddball and flanker stimuli elicit more salient and distinguishable neural responses than passive auditory oddball stimuli. For the comparison of method performance, we observe that the top-tier results are predominantly achieved by deep learning models trained from scratch. In contrast, existing foundation models for medical time series and EEG do not demonstrate clear performance advantages when fine-tuned from their released checkpoints. Moreover, manually extracted features rarely achieve top-tier performance, and extracted ERP-specific features consistently outperform generic EEG features in this stimulus type classification task.

\begin{table*}[t]
    \centering
    \scriptsize
    \caption{\textbf{ERP-Based Disease Detection Results.} ERP-based brain disease detection focuses on classifying neurological disorders, such as Parkinson’s disease or ADHD. \textcolor{myred}{\textbf{Top-1}}, \textcolor{myblue}{\underline{Top-2}}, and \textcolor{mygreen}{Top-3} results are highlighted in red, blue, and green.
    }
    \vspace{-2mm}
    \label{tab:erp_disease_results}
    \resizebox{\textwidth}{!}{
    \begin{tabular}{@{}ll|ccc|ccc|ccc@{}}
    \toprule

    \multicolumn{2}{l|}{\textbf{Datasets}}
    & \multicolumn{3}{c|}{ \makecell{\textbf{PD-SIM} \\ \textit{(55,921 Trials)} \\ \textit{(147 Subjects, 2 Classes)}} }
    & \multicolumn{3}{c|}{\makecell{\textbf{PD-ODD} \\ \textit{(34,464 Trials)} \\ \textit{(145 Subjects, 2 Classes)}} }
    & \multicolumn{3}{c}{\makecell{\textbf{ADHD-WMRI} \\ \textit{(21,832 Trials)} \\ \textit{(59 Subjects, 2 Classes)}}  }
    \\ \midrule

    \multicolumn{2}{l|}{\diagbox{\textbf{Methods}}{\textbf{Metrics}}} & \textbf{Acccuracy} & \textbf{F1 Score} & \textbf{AUROC} & \textbf{Acccuracy} & \textbf{F1 Score} & \textbf{AUROC} & \textbf{Acccuracy} & \textbf{F1 Score} & \textbf{AUROC} \\ \midrule

    \multicolumn{2}{l|}{\textbf{EEG Features}}  & 64.32\std{2.30} & 57.02\std{3.23}  & 64.41\std{4.44}  & 67.68\std{3.36} & 61.23\std{3.32}   & 70.01\std{3.42} & 58.43\std{3.07} & 55.76\std{3.61}  & 56.65\std{6.30}  \\
    \multicolumn{2}{l|}{\textbf{ERP Features}}  & 63.76\std{3.54} & 58.78\std{2.57}  & 66.41\std{3.71}  & 69.29\std{2.60} & 63.65\std{2.71}   & 72.05\std{2.50} & 59.79\std{4.79} & 57.40\std{5.93}  & 59.48\std{10.23}  \\

    \midrule
    \multicolumn{2}{l|}{\textbf{TCN}}  & 64.63\std{3.53} & 58.76\std{5.73}  & 64.92\std{5.32}  & 66.07\std{2.35} & 59.43\std{2.81}   & 68.50\std{2.28} & 62.74\std{2.46} & 60.57\std{2.82}  & 67.05\std{3.76}  \\
    \multicolumn{2}{l|}{\textbf{ModernTCN}}  & \textcolor{myred}{\textbf{71.98\std{1.82}}} & \textcolor{myblue}{\underline{65.79\std{2.58}}}  & \textcolor{myred}{\textbf{76.07\std{3.10}}}  & \textcolor{myblue}{\underline{71.74\std{1.47}}} & \textcolor{myblue}{\underline{66.72\std{0.73}}}   & \textcolor{myblue}{\underline{76.62\std{1.48}}} & 63.26\std{2.86} & 60.69\std{4.24}  & 66.79\std{4.88}  \\
    \multicolumn{2}{l|}{\textbf{TimesNet}}  & 63.09\std{3.86} & 54.44\std{4.99}  & 61.16\std{8.62}  & 68.15\std{2.83} & 60.45\std{2.37}   & 69.87\std{1.82} & \textcolor{myblue}{\underline{66.05\std{3.49}}} & \textcolor{myblue}{\underline{64.30\std{3.79}}}  & \textcolor{myblue}{\underline{72.12\std{5.04}}}  \\
    \multicolumn{2}{l|}{\textbf{PatchTST}}  & \textcolor{mygreen}{69.38\std{3.87}} & \textcolor{mygreen}{64.19\std{4.63}}  & \textcolor{mygreen}{73.32\std{4.77}}  & \textcolor{myred}{72.19\std{1.97}} & \textcolor{myred}{\textbf{67.37\std{1.15}}}   & \textcolor{myred}{\textbf{77.11\std{2.51}}} & 62.44\std{3.62} & 60.85\std{3.60}  & 66.46\std{5.03}  \\
    \multicolumn{2}{l|}{\textbf{iTransformer}}  & \textcolor{myblue}{\underline{71.38\std{2.09}}} & \textcolor{myred}{\textbf{66.79\std{2.39}}}  & \textcolor{myblue}{\underline{76.03\std{2.75}}}  & \textcolor{mygreen}{70.34\std{2.25}} & \textcolor{mygreen}{65.89\std{1.83}}   & \textcolor{mygreen}{75.06\std{1.77}} & 63.29\std{3.13} & 60.92\std{3.88}  & 66.77\std{5.35}  \\
    \multicolumn{2}{l|}{\textbf{Medformer}}  & 68.62\std{1.59} & 61.28\std{3.68}  & 70.95\std{2.71}  & 68.22\std{2.29} & 61.61\std{2.49}   & 72.10\std{2.81} & 64.48\std{2.60} & 61.47\std{3.25}  & 68.69\std{3.91}  \\
    \multicolumn{2}{l|}{\textbf{MedGNN}}  & 57.17\std{4.72} & 49.88\std{3.69}  & 56.94\std{6.91}  & 64.61\std{3.83} & 58.15\std{4.21}   & 68.59\std{3.61} & 62.70\std{3.98} & 59.97\std{5.71}  & 66.44\std{7.18}  \\
    \multicolumn{2}{l|}{\textbf{EEGNet}}  & 66.08\std{3.17} & 64.17\std{2.84}  & 71.63\std{3.45}  & 65.25\std{4.48} & 61.84\std{3.59}   & 70.46\std{5.57} & 52.70\std{8.00} & 44.77\std{5.24}  & 43.99\std{5.10}  \\
    \multicolumn{2}{l|}{\textbf{EEGInception}}  & 58.90\std{1.55} & 52.26\std{5.35}  & 57.52\std{8.47}  & 63.73\std{1.85} & 60.01\std{2.36}   & 65.97\std{1.88} & \textcolor{mygreen}{65.22\std{3.62}} & \textcolor{mygreen}{63.12\std{4.59}}  & \textcolor{mygreen}{69.60\std{5.73}}  \\
    \multicolumn{2}{l|}{\textbf{EEGConformer}}  & 60.69\std{4.64} & 56.80\std{5.22}  & 61.10\std{6.67}  & 65.44\std{1.40} & 62.88\std{1.56}   & 70.12\std{1.08} & \textcolor{myred}{\textbf{68.55\std{2.70}}} & \textcolor{myred}{\textbf{66.43\std{3.52}}}  & \textcolor{myred}{\textbf{74.55\std{4.68}}}  \\

    \midrule
    \multicolumn{2}{l|}{\textbf{BIOT}}  & 64.17\std{4.40} & 56.04\std{5.45}  & 62.37\std{6.41}  & 65.74\std{3.58} & 59.76\std{5.30}   & 69.03\std{3.47} & 54.29\std{5.62} & 53.05\std{5.73}  & 54.61\std{8.30}  \\
    \multicolumn{2}{l|}{\textbf{LaBraM}}  & 63.95\std{5.29} & 57.29\std{4.90}  & 63.47\std{6.11}  & 67.82\std{5.22} & 62.55\std{3.78}   & 71.62\std{7.46} & 63.82\std{9.56} & 62.11\std{9.61}  & 67.99\std{13.05}  \\
    \multicolumn{2}{l|}{\textbf{CBraMod}}  & 63.03\std{2.87} & 57.01\std{3.05}  & 64.73\std{3.72}  & 66.01\std{0.95} & 60.41\std{4.12}   & 71.06\std{2.71} & 61.98\std{5.24} & 59.96\std{5.40}  & 65.24\std{7.45}  \\
    \midrule

    \midrule
    \multicolumn{2}{l|}{\textbf{Datasets}}
    & \multicolumn{3}{c|}{\makecell{\textbf{SCPD} \\ \textit{(10,224 Trials)} \\ \textit{(56 Subjects, 2 Classes)}} }
    & \multicolumn{3}{c|}{\makecell{\textbf{RLPD} \\ \textit{(14,325 Trials)} \\ \textit{(56 Subjects, 2 Classes)}} }
    & \multicolumn{3}{c}{\makecell{\textbf{AOPD}  \\ \textit{(9,830 Trials)} \\ \textit{(50 Subjects, 2 Classes)}} }
    \\ \midrule

    \multicolumn{2}{l|}{\diagbox{\textbf{Methods}}{\textbf{Metrics}}} & \textbf{Acccuracy} & \textbf{F1 Score} & \textbf{AUROC} & \textbf{Acccuracy} & \textbf{F1 Score} & \textbf{AUROC} & \textbf{Acccuracy} & \textbf{F1 Score} & \textbf{AUROC} \\ \midrule

    \multicolumn{2}{l|}{\textbf{EEG Features}}  & 61.32\std{2.28} & 61.21\std{2.28}  & 66.63\std{3.19}  & 60.87\std{3.44} & 60.43\std{3.40}   & 62.70\std{4.09} & 61.83\std{3.32} & 61.43\std{3.55}  & 67.33\std{4.17}  \\
    \multicolumn{2}{l|}{\textbf{ERP Features}}  & 57.55\std{5.36} & 57.41\std{5.27}  & 61.93\std{8.01}  & 54.55\std{3.88} & 53.42\std{3.92}   & 58.89\std{6.35} & 61.43\std{2.73} & 60.82\std{3.08}  & 67.16\std{3.30}  \\

    \midrule
    \multicolumn{2}{l|}{\textbf{TCN}}  & 70.26\std{4.83} & 70.15\std{4.75}  & 77.31\std{6.78}  & \textcolor{mygreen}{68.67\std{2.67}} & \textcolor{mygreen}{68.40\std{2.59}}   & 73.23\std{6.78} & \textcolor{mygreen}{66.11\std{6.75}} & \textcolor{mygreen}{65.76\std{6.81}}  & 70.57\std{8.43}  \\
    \multicolumn{2}{l|}{\textbf{ModernTCN}}  & 67.85\std{5.39} & 67.55\std{5.37}  & 72.89\std{7.10}  & 61.04\std{5.63} & 60.75\std{5.46}   & 66.08\std{5.81} & 60.00\std{5.44} & 58.86\std{6.38}  & 62.06\std{8.09}  \\
    \multicolumn{2}{l|}{\textbf{TimesNet}}  & 71.44\std{6.30} & 71.15\std{6.34}  & 78.00\std{7.88}  & 67.64\std{7.49} & 67.36\std{7.46}   & 74.56\std{9.54} & 65.35\std{3.61} & 65.11\std{3.62}  & 71.53\std{4.52}  \\
    \multicolumn{2}{l|}{\textbf{PatchTST}}  & 63.37\std{4.36} & 63.15\std{4.39}  & 68.80\std{5.95}  & 61.07\std{7.23} & 60.35\std{7.20}   & 64.87\std{8.99} & 57.63\std{6.00} & 56.92\std{6.51}  & 60.39\std{8.29}  \\
    \multicolumn{2}{l|}{\textbf{iTransformer}}  & 66.83\std{5.65} & 66.51\std{5.83}  & 71.91\std{7.25}  & 59.36\std{4.23} & 59.07\std{4.14}   & 62.90\std{5.05} &  59.56\std{6.43} & 58.54\std{7.36}  & 61.05\std{9.63}  \\
    \multicolumn{2}{l|}{\textbf{Medformer}}  & 65.22\std{6.80} & 64.93\std{6.96}  & 71.91\std{7.87}  & 63.45\std{6.43} & 63.18\std{6.20}   & 69.80\std{8.35} & 64.94\std{6.59} & 64.59\std{6.60}  & 70.56\std{8.78}  \\
    \multicolumn{2}{l|}{\textbf{MedGNN}}  & 66.13\std{9.71} & 65.73\std{9.97}  & 72.34\std{12.06}  & 64.68\std{6.65} & 64.36\std{6.80}   & 72.76\std{8.73} & 62.93\std{3.93} & 62.48\std{4.00}  & 69.29\std{5.56}  \\
    \multicolumn{2}{l|}{\textbf{EEGNet}}  & 64.78\std{4.61} & 63.57\std{4.96}  & 68.41\std{5.85}  & 63.18\std{1.77} & 62.53\std{1.61}   & 67.12\std{2.79} & 62.51\std{7.17} & 60.41\std{8.64}  & 64.92\std{9.70}  \\
    \multicolumn{2}{l|}{\textbf{EEGInception}}  & \textcolor{myred}{\textbf{74.48\std{5.53}}} & \textcolor{myred}{\textbf{74.33\std{5.42}}}  & \textcolor{myred}{\textbf{82.79\std{6.36}}}  & \textcolor{myblue}{\underline{70.76\std{4.43}}} & \textcolor{myblue}{\underline{70.57\std{4.42}}}   & \textcolor{myblue}{\underline{77.32\std{6.43}}} & \textcolor{myred}{\textbf{69.51\std{6.59}}} & \textcolor{myred}{\textbf{69.26\std{6.52}}}  & \textcolor{myred}{\textbf{77.32\std{8.81}}}  \\
    \multicolumn{2}{l|}{\textbf{EEGConformer}}  & \textcolor{myblue}{\underline{72.35\std{5.39}}} & \textcolor{myblue}{\underline{72.04\std{5.43}}}  & \textcolor{mygreen}{79.43\std{6.74}}  & 66.09\std{4.05} & 65.28\std{4.16}   & \textcolor{mygreen}{75.73\std{6.46}} & \textcolor{myblue}{\underline{68.25\std{9.26}}} & \textcolor{myblue}{\underline{68.00\std{9.26}}}  & \textcolor{myblue}{\underline{74.43\std{11.74}}}  \\

    \midrule
    \multicolumn{2}{l|}{\textbf{BIOT}}  & 70.00\std{4.05} & 69.42\std{4.61}  & 76.71\std{6.79}  & 65.72\std{9.04} & 65.09\std{8.85}   & 69.29\std{12.06} & 62.15\std{8.73} & 61.57\std{8.80}  & 66.51\std{10.06}  \\
    \multicolumn{2}{l|}{\textbf{LaBraM}}  & \textcolor{mygreen}{72.27\std{4.17}} & \textcolor{mygreen}{71.71\std{4.30}}  & \textcolor{myblue}{\underline{79.46\std{6.52}}}  & \textcolor{myred}{\textbf{74.47\std{6.34}}} & \textcolor{myred}{\textbf{73.87\std{6.94}}}   & \textcolor{myred}{\textbf{84.82\std{5.72}}} & 62.05\std{9.25} & 61.55\std{9.20}  & 67.40\std{13.80}  \\
    \multicolumn{2}{l|}{\textbf{CBraMod}}  & 67.33\std{3.45} & 66.63\std{3.89}  & 73.34\std{8.45}  & 63.28\std{9.41} & 62.65\std{9.14}   & 66.26\std{12.52} & 65.99\std{3.78} & 65.28\std{3.88}  & \textcolor{mygreen}{71.43\std{4.80}}  \\

    \bottomrule
    \end{tabular}
    }
\vspace{-2mm}
\end{table*}

\begin{figure*}[t]
    \centering
    \includegraphics[width=1.0\linewidth]{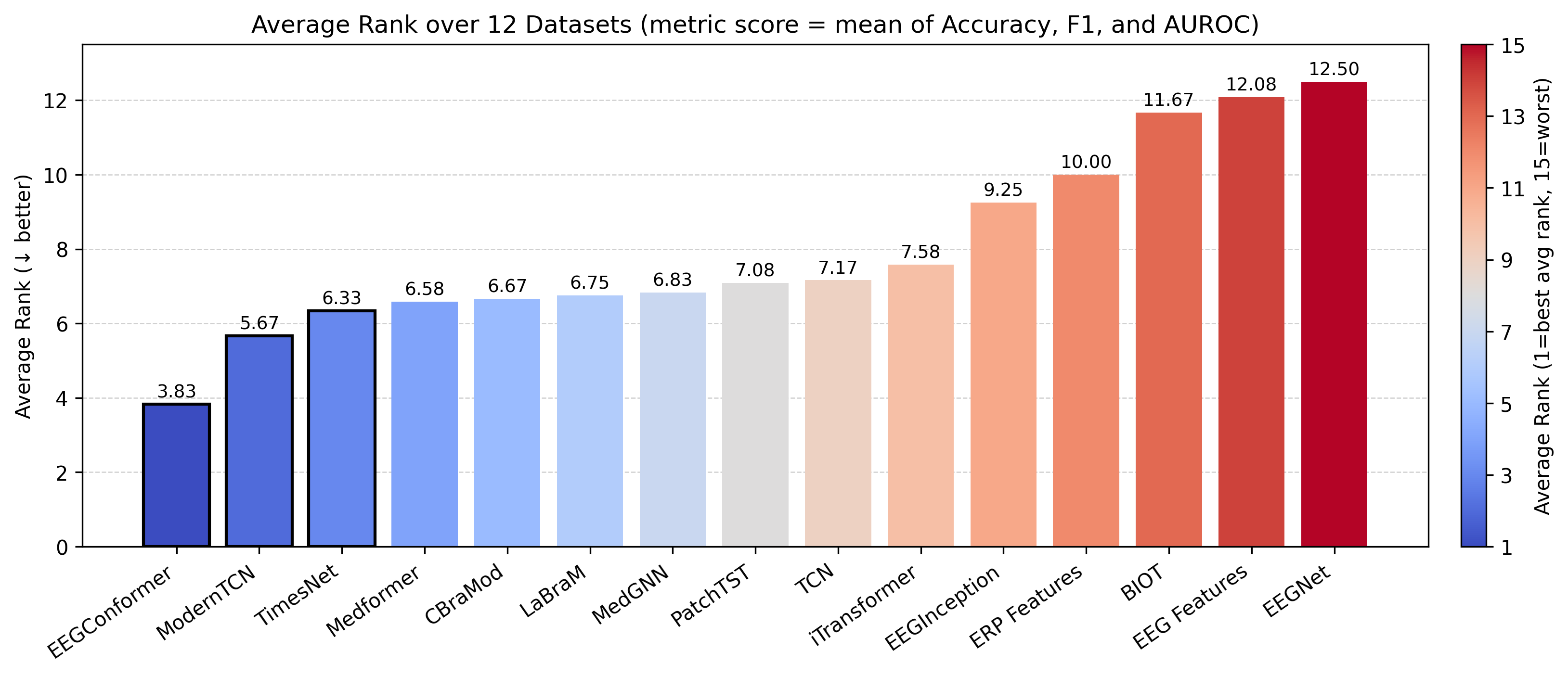}
    \caption{\textbf{Overall Average Rank.} Average performance rank of 15 methods across all 12 datasets. For example, the value 3.83 for EEGConformer indicates an average rank across all 12 datasets.
    }
    \label{fig:avg_rank_bar_heatmap}
    \vspace{-3mm}
\end{figure*}

\begin{table*}[t]
    \centering
    \caption{\textbf{Method Ranks Across 12 Datasets.} For each dataset, we first average Accuracy, F1 score, and AUROC to obtain a dataset-level score for each method, and then rank all methods based on this score. The last column reports the overall average rank across all 12 datasets. Lower rank indicates better performance. For each column, \textcolor{myred}{\textbf{Top-1}}, \textcolor{myblue}{\underline{Top-2}}, and \textcolor{mygreen}{Top-3} results are highlighted in red, blue, and green, respectively.}
    \vspace{-2mm}
    \label{tab:avg_rank}
    \resizebox{\textwidth}{!}{%
    \begin{tabular}{@{}l|cccccc|cccccc|c@{}}
    \toprule
    \multirow{3}{*}{\textbf{Method}}
    & \multicolumn{6}{c|}{\textbf{Stimulus Classification}}
    & \multicolumn{6}{c|}{\textbf{Disease Detection}}
    & \multirow{3}{*}{\makecell{\textbf{Overall} \\ \textbf{Average} \\ \textbf{Rank}}} \\
    \cmidrule(lr){2-7} \cmidrule(lr){8-13}
    & \makecell{\textbf{CESCA-} \\ \textbf{AODD}}
    & \makecell{\textbf{CESCA-} \\ \textbf{VODD}}
    & \makecell{\textbf{CESCA-} \\ \textbf{FLANKER}}
    & \makecell{\textbf{mTBI-} \\ \textbf{ODD}}
    & \makecell{\textbf{NSERP-} \\ \textbf{MSIT}}
    & \makecell{\textbf{NSERP-} \\ \textbf{ODD}}
    & \makecell{\textbf{PD-} \\ \textbf{SIM}}
    & \makecell{\textbf{PD-} \\ \textbf{ODD}}
    & \makecell{\textbf{ADHD-} \\ \textbf{WMRI}}
    & \textbf{SCPD}
    & \textbf{RLPD}
    & \textbf{AOPD}
    & \\
    \midrule
    \textbf{EEGConformer} & 6.00 & \textcolor{myred}{\textbf{1.00}} & \textcolor{mygreen}{3.00} & \textcolor{myred}{\textbf{1.00}} & \textcolor{myblue}{\underline{2.00}} & \textcolor{myred}{\textbf{1.00}} & 13.00 & 9.00 & \textcolor{myred}{\textbf{1.00}} & \textcolor{myblue}{\underline{2.00}} & 5.00 & \textcolor{myblue}{\underline{2.00}} & \textcolor{myred}{\textbf{3.83}} \\
    \textbf{ModernTCN} & 7.00 & 5.00 & \textcolor{myblue}{\underline{2.00}} & \textcolor{myblue}{\underline{2.00}} & 5.00 & 5.00 & \textcolor{myblue}{\underline{2.00}} & \textcolor{myblue}{\underline{2.00}} & 7.00 & 7.00 & 11.00 & 13.00 & \textcolor{myblue}{\underline{5.67}} \\
    \textbf{TimesNet} & 5.00 & 6.00 & 11.00 & 7.00 & 6.00 & 6.00 & 12.00 & 8.00 & \textcolor{myblue}{\underline{2.00}} & 4.00 & 4.00 & 5.00 & \textcolor{mygreen}{6.33} \\
    \textbf{Medformer} & \textcolor{myred}{\textbf{1.00}} & 9.00 & 8.00 & 8.00 & 4.00 & 9.00 & 5.00 & 6.00 & 4.00 & 11.00 & 8.00 & 6.00 & 6.58 \\
    \textbf{CBraMod} & 8.00 & 8.00 & 4.00 & 5.00 & \textcolor{myred}{\textbf{1.00}} & \textcolor{myblue}{\underline{2.00}} & 9.00 & 11.00 & 11.00 & 8.00 & 10.00 & \textcolor{mygreen}{3.00} & 6.67 \\
    \textbf{LaBraM} & \textcolor{myblue}{\underline{2.00}} & 11.00 & 9.00 & 10.00 & 9.00 & 8.00 & 10.00 & 5.00 & 5.00 & \textcolor{mygreen}{3.00} & \textcolor{myred}{\textbf{1.00}} & 8.00 & 6.75 \\
    \textbf{MedGNN} & 4.00 & 4.00 & \textcolor{myred}{\textbf{1.00}} & 4.00 & \textcolor{mygreen}{3.00} & 4.00 & 15.00 & 14.00 & 10.00 & 10.00 & 6.00 & 7.00 & 6.83 \\
    \textbf{PatchTST} & \textcolor{mygreen}{3.00} & \textcolor{myblue}{\underline{2.00}} & 6.00 & 6.00 & 8.00 & 7.00 & \textcolor{mygreen}{3.00} & \textcolor{myred}{\textbf{1.00}} & 9.00 & 13.00 & 12.00 & 15.00 & 7.08 \\
    \textbf{TCN} & 10.00 & \textcolor{mygreen}{3.00} & 10.00 & 9.00 & 11.00 & \textcolor{mygreen}{3.00} & 7.00 & 13.00 & 8.00 & 5.00 & \textcolor{mygreen}{3.00} & 4.00 & 7.17 \\
    \textbf{iTransformer} & 9.00 & 7.00 & 7.00 & \textcolor{mygreen}{3.00} & 7.00 & 11.00 & \textcolor{myred}{\textbf{1.00}} & \textcolor{mygreen}{3.00} & 6.00 & 9.00 & 14.00 & 14.00 & 7.58 \\
    \textbf{EEGInception} & 15.00 & 12.00 & 12.00 & 12.00 & 12.00 & 12.00 & 14.00 & 15.00 & \textcolor{mygreen}{3.00} & \textcolor{myred}{\textbf{1.00}} & \textcolor{myblue}{\underline{2.00}} & \textcolor{myred}{\textbf{1.00}} & 9.25 \\
    \textbf{ERP Features} & 11.00 & 10.00 & 5.00 & 11.00 & 10.00 & 10.00 & 6.00 & 4.00 & 12.00 & 15.00 & 15.00 & 11.00 & 10.00 \\
    \textbf{BIOT} & 14.00 & 13.00 & 14.00 & 13.00 & 13.00 & 13.00 & 11.00 & 12.00 & 14.00 & 6.00 & 7.00 & 10.00 & 11.67 \\
    \textbf{EEG Features} & 12.00 & 14.00 & 13.00 & 14.00 & 14.00 & 14.00 & 8.00 & 7.00 & 13.00 & 14.00 & 13.00 & 9.00 & 12.08 \\
    \textbf{EEGNet} & 13.00 & 15.00 & 15.00 & 15.00 & 15.00 & 15.00 & 4.00 & 10.00 & 15.00 & 12.00 & 9.00 & 12.00 & 12.50 \\
    \bottomrule
    \end{tabular}%
    }
    \vspace{-4mm}
\end{table*}

\subsection{ERP-Based Disease Detection Results}
\label{sub:erp_disease_results}

We then investigate ERP-based disease detection performance, with results summarized in Tables~\ref{tab:erp_disease_results}. All experiments are conducted under a subject-independent evaluation setup, aiming to identify disease-specific ERP features that generalize across subjects for real-world disease detection applications. Across the six datasets (PD-SIM, PD-ODD, ADHD-WMRI, SCPD, RLPD, and AOPD), the best method achieves F1 scores of 66.79\%, 67.37\%, 66.43\%, 74.33\%, 73.87\%, and 69.26\%, respectively. Overall, no single ERP paradigm consistently demonstrates a clear advantage for disease detection. Notably, the three smaller Parkinson’s disease datasets, SCPD, RLPD, and AOPD, achieve relatively higher performance than the two larger datasets, PD-SIM and PD-ODD. This trend may be attributed to the more balanced data distributions in the smaller datasets, which contain exactly equal numbers of Parkinson’s disease and healthy control subjects. For method comparison, top-tier performance is primarily achieved by deep learning models trained from scratch and by models fine-tuned from foundation model checkpoints. Two methods based on manually extracted features never achieve top-tier results. Besides, unlike ERP stimulus-type classification in the previous subsection, ERP-specific features do not consistently outperform generic EEG features in disease detection. In several datasets, ERP-specific features even perform worse, suggesting that the disease-related information in these datasets may be less tightly aligned with canonical ERP components.

\subsection{Overall Performance Analysis}
\label{sub:overall_results}

To provide a comprehensive comparison of methods across multiple tasks and evaluation metrics, we further analyze their overall performance across 12 datasets. For each dataset, we first compute a dataset-level score for method $j$ by averaging its Accuracy, F1 score, and AUROC:
\begin{equation}
    s_{i,j} = \frac{\mathrm{Acc}_{i,j} + \mathrm{F1}_{i,j} + \mathrm{AUROC}_{i,j}}{3},
\end{equation}
where $s_{i,j}$ denotes the dataset-level score of method $j$ on dataset $i$. Based on these scores, all methods are ranked within each dataset. Let $r_{i,j}$ denote the rank of method $j$ on dataset $i$. The overall average rank of method $j$ across all datasets is then computed as
\begin{equation}
    R_j = \frac{1}{N} \sum_{i=1}^{N} r_{i,j},
\end{equation}
where $N$ is the number of datasets and $k$ is the number of compared methods. Figure~\ref{fig:avg_rank_bar_heatmap} presents the overall average rank of each method across 12 datasets. Table~\ref{tab:avg_rank} provides the detailed ranks of each method on each dataset as complementary information. Each value in the table (except for the last column) corresponds to $r_{i,j}$, while the last column reports the overall average rank $R_j$.

We observe that supervised deep learning models trained from scratch consistently outperform other methods in overall performance. Among all evaluated approaches, \textbf{EEGConformer} achieves the lowest average rank of 3.83, indicating the most stable and competitive performance across diverse tasks and datasets. Following EEGConformer, ModernTCN, TimesNet, Medformer, and CBraMod constitute the top-tier group. Notably, only CBraMod falls within the foundation model category, whereas the remaining methods are trained from scratch. This suggests that existing EEG foundation models do not demonstrate clear performance advantages over supervised deep learning models trained from scratch. In addition, manual feature-based methods consistently rank in the lowest performance tier, while most deep learning approaches easily outperform them. Between the two handcrafted feature methods, ERP-specific features perform relatively better than generic EEG features. These results highlight the limitations of handcrafted statistical and frequency-domain features, which struggle to maintain robust performance across diverse ERP paradigms and disease conditions, particularly under subject-independent evaluation. This suggests that, beyond interpretability or neuroscience-oriented analysis, manual feature extraction offers limited advantages in real-world applications where end-to-end performance is the primary objective.

\begin{table}[h]
    \centering
    \small
    \caption{Pairwise comparisons between EEGConformer and other methods using Holm-corrected paired $t$-tests. Here, $\Delta$F1 denotes the mean F1 difference ($\Delta$F1 = EEGConformer $-$ method), and the 95\% confidence interval (CI) is computed across all dataset-seed evaluation blocks.}
    \label{tab:p_values}

    \begin{tabular}{lccc}
    \toprule
    \textbf{Method} & \textbf{Mean $\Delta$F1} & \textbf{95\% CI} & \textbf{$p$-value} \\
    \midrule
    LaBraM & 0.0142 & [-0.0005, 0.0290] & 0.0585 \\
    ModernTCN & 0.0194 & [0.0052, 0.0335] & 0.0167 \\
    iTransformer & 0.0239 & [0.0086, 0.0391] & 0.0083 \\
    PatchTST & 0.0252 & [0.0101, 0.0404] & 0.0059 \\
    TimesNet & 0.0171 & [0.0071, 0.0271] & 0.0058 \\
    Medformer & 0.0194 & [0.0083, 0.0304] & 0.0051 \\
    CBraMod & 0.0227 & [0.0111, 0.0342] & 0.0016 \\
    TCN & 0.0208 & [0.0113, 0.0302] & $<0.001$ \\
    MedGNN & 0.0318 & [0.0192, 0.0444] & $<0.001$ \\
    EEGInception & 0.0507 & [0.0334, 0.0680] & $<0.001$ \\
    ERP Features & 0.0572 & [0.0404, 0.0740] & $<0.001$ \\
    EEGNet & 0.1344 & [0.1018, 0.1669] & $<0.001$ \\
    BIOT & 0.0921 & [0.0697, 0.1144] & $<0.001$ \\
    EEG Features & 0.1212 & [0.0931, 0.1492] & $<0.001$ \\
    \bottomrule
    \end{tabular}

    \vspace{-5mm}
\end{table}

\begin{table*}[t]
    \centering
    \caption{\textbf{Embedding Method Comparison.} Three commonly used token embedding methods in existing EEG Transformer models. Patch lengths of 25, 50, and 100 are used for multi-variate and uni-variate embeddings to eliminate performance change caused by patch length. All other Transformer components are kept identical. Results are reported in the F1 score.
    }
    \vspace{-2mm}
    \label{tab:embedding_comparison}
    \resizebox{1.0\textwidth}{!}{
    \begin{tabular}{@{}ll|ccc|ccc|c@{}}
    \toprule

    \multicolumn{2}{l|}{\diagbox{\textbf{Datasets}}{\textbf{Methods}}}  & \makecell{\textbf{Multi-Variate} \\ \textit{(25)} } & \makecell{\textbf{Multi-Variate} \\ \textit{(50)} } & \makecell{\textbf{Multi-Variate} \\ \textit{(100)} }  & \makecell{\textbf{Uni-Variate} \\ \textit{(25)} } & \makecell{\textbf{Uni-Variate} \\ \textit{(50)} } & \makecell{\textbf{Uni-Variate} \\ \textit{(100)} } & \makecell{\textbf{Whole-Variate} \\ \textit{(---)} }
    \\ \midrule

    \multicolumn{2}{l|}{\textbf{CESCA-AODD}}  &  53.26\std{0.70} & 53.41\std{0.59}  & 53.15\std{0.69}  &  53.91\std{0.97} & 53.55\std{0.72}  & \textbf{54.02\std{0.73}}  & 53.35\std{0.81}  \\
    \multicolumn{2}{l|}{\textbf{CESCA-VODD}}  &  67.00\std{1.62} & 66.10\std{1.98}  & 67.27\std{1.33}  &  67.47\std{1.83} & \textbf{68.27\std{1.68}}  & 67.71\std{1.77}  & 65.66\std{1.33}  \\
    \multicolumn{2}{l|}{\textbf{CESCA-FLANKER}}  &  63.21\std{0.84} & 63.76\std{0.99}  & 63.38\std{0.83}  &  63.73\std{0.87} & \textbf{63.87\std{1.13}}  & 63.81\std{0.82}  & 63.35\std{1.39}  \\
    \multicolumn{2}{l|}{\textbf{mTBI-ODD}}  &  63.41\std{1.62} & 63.61\std{1.61}  & 63.18\std{1.79}  &  63.88\std{1.61} & 64.15\std{1.91}  & \textbf{64.51\std{1.83}}  & 63.98\std{1.67}  \\
    \multicolumn{2}{l|}{\textbf{NSERP-MSIT}}  &  37.34\std{2.21} & 36.87\std{2.00}  & 36.62\std{2.42}  &  \textbf{37.54\std{2.31}} & 36.65\std{2.51}  & 36.87\std{2.85}  & 35.69\std{2.37}  \\
    \multicolumn{2}{l|}{\textbf{NSERP-ODD}}  &  \textbf{65.36\std{2.32}} & 64.88\std{1.77}  & 63.62\std{2.21}  &  65.21\std{1.72} & 63.96\std{2.55}  & 63.59\std{2.61}  & 62.39\std{2.69}  \\

    \midrule
    \multicolumn{2}{l|}{\textbf{PD-SIM}}  &  58.16\std{5.05} & 63.46\std{2.14}  & 64.33\std{1.72}  &  64.41\std{4.28} & 66.59\std{2.27}  & \textbf{68.41\std{1.73}}  & 67.14\std{2.55}  \\
    \multicolumn{2}{l|}{\textbf{PD-ODD}}  &  62.45\std{3.19} & 63.03\std{2.16}  & 64.17\std{2.08}  &  68.03\std{0.97} & \textbf{68.85\std{1.30}}  & 68.50\std{0.44}  & 66.13\std{1.58}  \\
    \multicolumn{2}{l|}{\textbf{ADHD-WMRI}}  &  \textbf{63.77\std{5.06}} & 62.28\std{3.82}  & 61.30\std{2.88}  &  61.03\std{3.90} & 60.17\std{3.94}  & 60.53\std{3.15}  & 60.84\std{3.37}  \\
    \multicolumn{2}{l|}{\textbf{SCPD}}  &  67.84\std{6.87} & 65.90\std{6.48}  & 62.18\std{7.83}  &  65.96\std{5.40} & \textbf{68.05\std{5.66}}  & 67.91\std{5.04}  & 65.44\std{5.03}  \\
    \multicolumn{2}{l|}{\textbf{RLPD}}  &  \textbf{64.47\std{4.74}} & 61.19\std{5.68}  & 62.51\std{4.36}  &  62.56\std{5.18} & 61.47\std{5.46}  & 60.73\std{4.92}  & 59.48\std{5.00}  \\
    \multicolumn{2}{l|}{\textbf{AOPD}}  &  \textbf{65.29\std{6.78}} & 63.49\std{7.32}  & 59.96\std{4.39}  &  58.70\std{4.80} & 59.49\std{7.46}  & 59.43\std{6.76}  & 59.19\std{6.45}  \\

    \bottomrule
    \end{tabular}
    }
\vspace{-2mm}
\end{table*}

\subsection{Statistical Significance Analysis}
\label{sub:statistical_analysis}

We first performed a repeated-measures ANOVA across all compared methods using F1 scores. Since our benchmark adopts a Monte Carlo subject-independent evaluation setup, different seeds correspond to different subject partitions for the train/validation/test split. Therefore, each dataset-seed pair is treated as a repeated evaluation block, with method as the within-block factor. When the ANOVA indicated significant differences, we conducted post-hoc pairwise comparisons between the best-ranked method, EEGConformer, and all other methods using paired $t$-tests with Holm correction.

The repeated-measures ANOVA revealed significant performance differences among the compared methods across the 12 datasets ($p < 0.001$). Therefore, the null hypothesis that all methods achieve equivalent performance is rejected, and post-hoc paired $t$-tests are conducted to assess pairwise statistical significance. The results are presented in Table~\ref{tab:p_values}, which reports the mean F1 difference ($\Delta$F1 = EEGConformer $-$ method), the corresponding 95\% confidence interval, and the Holm-corrected $p$-value for each pairwise comparison. EEGConformer significantly outperforms most compared methods, while LaBraM shows only a borderline difference, as its confidence interval slightly overlaps zero. Overall, these results provide strong statistical evidence supporting the superiority of EEGConformer for ERP classification, consistent with the average ranking results in Table~\ref{tab:avg_rank}.

\subsection{Embedding Method Comparison for Transformer}
\label{sub:embedding_comparison}

We compare three commonly used token embedding methods in existing EEG Transformer models to examine their suitability for ERP data classification. Patch lengths of 25, 50, and 100 are used for multi-variate and uni-variate embeddings to eliminate performance fluctuation caused by patch length. All other components, including the self-attention modules, feedforward networks, and layer normalization layers, are kept identical across different embedding methods. 

All results are reported in terms of F1 score in Table~\ref{tab:embedding_comparison}. The uni-variate embedding method achieves the best performance on 8 out of 12 datasets, while the multi-variate embedding method achieves the best results on the remaining 4 datasets. In contrast, the whole-variate embedding method does not achieve top performance on any dataset. These results demonstrate the advantage of uni-variate token embedding for ERP classification and suggest promising start directions for designing ERP-specific Transformer backbones. Moreover, we observe that finer self-attention with shorter patch lengths yields advantages only in the multi-variate token embedding setting, where all 4 best-performing cases occur with a patch length of 25. In contrast, uni-variate token embedding does not consistently achieve the best performance at a specific patch length, and the optimal patch length varies across datasets. This observation suggests that appropriate patch-length selection remains important when using uni-variate token embeddings.

\section{Discussion}
\label{sec:discussion}

Overall, performance on both stimulus-type classification and brain disease detection remains relatively limited for real-world deployment. This limitation may stem from a combination of dataset characteristics and methodological choices. Although the number of trials in existing ERP datasets is often sufficient for traditional manual feature-based analysis, it may still be inadequate for training data-hungry deep learning models. In addition, dataset quality factors, such as noise contamination and class imbalance, can further constrain achievable performance, as discussed in section~\ref{sub:erp_disease_results} by comparing Parkinson’s disease detection performance across different datasets.

From a method perspective, EEGConformer, a model proposed several years ago, achieves the best average ranking in overall performance, outperforming more recent approaches, including foundation models. This is notable given that foundation models typically contain more parameters and require substantially longer training time, even when fine-tuned from pre-trained checkpoints. This observation suggests that recent advances in general EEG representation learning do not directly translate to the ERP sub-domain, highlighting the need for backbone architectures specifically tailored to ERP data characteristics. The comparison of embedding methods (Table~\ref{tab:embedding_comparison}) for the Transformer provides a useful starting point for designing ERP-specific Transformer architectures. The results suggest that uni-variate embedding may be a better choice for token embedding, although the optimal patch length remains challenging to determine. Future work on exploring multi-scale learning strategies may be a solution, similar to those used in Medformer~\cite{wang2024medformer}.

Moreover, the lack of a clear advantage of existing foundation models over fully supervised training suggests that ERP-centric pre-training may also be necessary. Most current EEG foundation models are primarily pre-trained on spontaneous EEG corpora. For example, BIOT was pre-trained on the resting-state EEG dataset PREST~\cite{yang2023biot}; LaBraM was pre-trained on 16 datasets totaling 2,500 hours, among which only 4 datasets (81.16 hours) contain ERP recordings; and CBraMod was trained on TUEG~\cite{obeid2016temple}, which consists entirely of spontaneous EEG data. Furthermore, none of these models incorporates ERP-specific preprocessing procedures or pre-training strategies. However, ERP data are organized as time-locked trials with stimulus-specific temporal dynamics, which differ fundamentally from spontaneous EEG. These domain gaps suggest that ERP-oriented datasets, preprocessing pipelines, and pre-training methods are likely crucial for improving downstream ERP classification performance.

\section{Conclusion}
\label{sec:conclusion}

In this paper, we conduct a comprehensive benchmark study of two ERP tasks: ERP stimulus-type classification and ERP-based brain disease detection, across 12 datasets. We compare 15 methods, including two manual feature-extraction approaches, ten supervised deep learning methods, and three EEG foundation models with pre-trained weights. In addition, to investigate the most suitable token embedding strategy for Transformer-based ERP classification, we conduct a controlled comparison of three commonly used patch embedding methods, aiming to inform the design of ERP-specific Transformer architectures. Based on the results from 36 evaluations, we summarize four key takeaways from this benchmark study. First, modern deep learning methods consistently outperform manual feature extraction approaches on ERP tasks. Second, existing EEG foundation models do not exhibit clear performance advantages over supervised deep learning models trained from scratch. Third, EEGConformer achieves the most competitive overall performance among the evaluated methods for ERP classification. Fourth, uni-variate patch embedding demonstrates the strongest performance among the examined embedding strategies for ERP-specific Transformers. We hope that this benchmark study serves as a useful reference for method selection in ERP analysis and inspires future research on designing more effective ERP-oriented models for advanced ERP analysis and representation learning.

\bibliographystyle{IEEEbib}  
\bibliography{refs}



\end{document}